\documentclass[pre,twocolumn,twoside,byrevtex,superscriptaddress,floatfix,nofootinbib]{revtex4-1}

\usepackage{libs}
\usepackage{settings}

\setboolean{twocolswitch}{true}
\begin{document}

\title{\protect The growing amplification of social media:\\
Measuring temporal and social contagion dynamics\\
for over 150 languages on Twitter for 2009--2020 

%% Sociolinguistic echo chamber: Temporal dynamics of over 150 languages on Twitter from 2008--2020 
%% \newline
%% \underline{OR}\\
%% \newline
%% Sociolinguistic Wildfire: Temporal dynamics of over 150 languages on Twitter from 2008--2020 
%% \newline
%% \underline{OR}
%% \mbox{}\\
%% Temporal dynamics of sociolinguistic contagion of over 150 languages on Twitter from 2008--2020 
%% \\
%% 
%% \newline
%% Over 150 languages
%% from 2008--2020
%% Relative volume
%% Contagion
%% Echo chamber
%% sociolinguistic wildfire

%Temporal dynamics of 170 languages on Twitter from 2008-2020
%Exploring a decade of language evolution for over 170 languages on Twitter
}
\author{
\firstname{Thayer}
\surname{Alshaabi}
}
\email{thayer.alshaabi@uvm.edu}
\affiliation{
  Vermont Complex Systems Center,
  University of Vermont,
  Burlington, VT 05405.
  }
\affiliation{
  Computational Story Lab,
  University of Vermont,
  Burlington, VT 05405.
  }
 \affiliation{
  Department of Computer Science,\\
  University of Vermont,
  Burlington, VT 05405.
  }

\author{
    \firstname{David R.}
    \surname{Dewhurst}
}
\affiliation{
  Vermont Complex Systems Center,
  University of Vermont,
  Burlington, VT 05405.
  }
\affiliation{
  Computational Story Lab,
  University of Vermont,
  Burlington, VT 05405.
  }
\affiliation{
  Charles River Analytics, 
  Cambridge, MA 02138.
  }

\author{ 
  \firstname{Joshua R.}
  \surname{Minot}
}
\affiliation{
  Vermont Complex Systems Center,
  University of Vermont,
  Burlington, VT 05405.
  }
\affiliation{
  Computational Story Lab,
  University of Vermont,
  Burlington, VT 05405.
  }
  
\author{
  \firstname{Michael V.}
  \surname{Arnold}
}
\affiliation{
  Vermont Complex Systems Center,
  University of Vermont,
  Burlington, VT 05405.
  }
\affiliation{
  Computational Story Lab,
  University of Vermont,
  Burlington, VT 05405.
  }
  
\author{
  \firstname{Jane L.}
  \surname{Adams}
}
\affiliation{
  Vermont Complex Systems Center,
  University of Vermont,
  Burlington, VT 05405.
  }
\affiliation{
  Computational Story Lab,
  University of Vermont,
  Burlington, VT 05405.
  }

\author{
  \firstname{Christopher M.}
  \surname{Danforth}
}
%% \email{chris.danforth@uvm.edu}
\affiliation{
  Vermont Complex Systems Center,
  University of Vermont,
  Burlington, VT 05405.
  }
\affiliation{
  Computational Story Lab,
  University of Vermont,
  Burlington, VT 05405.
  }
\affiliation{
  Department of Mathematics \& Statistics,\\
  University of Vermont,
  Burlington, VT 05405.
  }

\author{
  \firstname{Peter Sheridan}
  \surname{Dodds}
}
\email{peter.dodds@uvm.edu}
\affiliation{
  Vermont Complex Systems Center,
  University of Vermont,
  Burlington, VT 05405.
  }
\affiliation{
  Computational Story Lab,
  University of Vermont,
  Burlington, VT 05405.
  }
\affiliation{
  Department of Computer Science,\\
  University of Vermont,
  Burlington, VT 05405.
  }

\date{\today}

\begin{abstract}
  \protect
  Working from a dataset of 118 billion messages running from the start of 2009 to the end of 2019, 
we identify and explore the relative daily use of over 150 languages on Twitter.
We find that eight languages comprise 80\% of all tweets, 
with English, Japanese, Spanish, Arabic, and Portuguese being the most dominant.
To quantify social spreading in each language over time, we compute the ‘contagion ratio': The balance of retweets to organic messages. 
We find that for the most common languages on Twitter there is a growing tendency, though not universal, to retweet rather than share new content.
By the end of 2019, the contagion ratios for half of the top 30 languages, including English and Spanish, had reached above 1---the naive contagion threshold.
In 2019, 
the top 5 languages with
the highest average daily ratios were, in order,
Thai (7.3), 
Hindi, 
Tamil,
Urdu,
and
Catalan,
while the bottom 5 were
Russian,
Swedish,
Esperanto,
Cebuano,
and Finnish (0.26).
Further, we show that over time,
the contagion ratios for
most common languages are growing 
more strongly than those of rare languages.
 
\end{abstract}


%% 89.65.-s	Social and economic systems
%% 89.75.Da	Systems obeying scaling laws
%% 89.75.Fb	Structures and organization in complex systems
%% 89.75.-k	Complex systems (for complex chemical systems, see 82.40.Qt; for biological complexity, see 87.18.-h)

\maketitle   

%%%%%%%%% end of author(s), address(es) plus abstract

%% add sections here...
%%%%%%%%%%%%%%%%%%%%%%%%%%%%%%%%%%%%%%%%%%%%%%%%%%%%%%%%%%%
\section{Introduction}
\label{sec:introduction} 
%% Suggest saving for thesis introduction:
%% 
%% Quotations are often used for numerous number of reasons. 
%% For example, we share quotes with our friends and colleagues to inspire from. 
%% Academics cite others standing on the shoulders of giants and reshaping our future.
%% People may quote scientists and politicians to acknowledge their contributions. 
%% Fans post and re-post information about their favorite movies, and their favorite %% bands
%% to share their stories. 
%% The exercise of 
%% quoting---repeating someone's else words (e.g., retweeting in the context of 
%% Twitter)---is in itself a spreading event de novo. 
%% Different languages, however, will have different cultural norms around this %% profoundly 
%% ancient phenomenon that dates back to centuries ago. 
%% Recent development of social media 
%% has enabled this contagious behaviour to be observed on a larger scale. 
%% Many studies address theoretical models of social contagion but without adequate data---in as many different written languages, the dominant mode of online human communication, as is practical---these theoretical modeling studies cannot be effectively confronted with reality.

Users of social media are presented with a choice:
post nothing at all; post something original; 
or re-post (``retweet'' in the case of Twitter) an existing post.
The simple amplifying mechanism of reposting encodes a unique digital and behavioral aspect of social contagion, with increasingly important ramifications
as interactions and conversations on social media platforms such as Twitter
tend to mirror the dynamics of major global and local events~\cite{suh2010want,boyd2010tweet,nagarajan2010qualitative,hodas2012visibility}.

Previous studies have explored the role of retweeting in the social contagion literature,
though the vast majority of this research is limited to either a given language 
(e.g., English tweets) 
or a short period~\cite{suh2010want,boyd2010tweet,harrigan2012influentials,hodas2014simple}. 
Here,
drawing on a 10\% random sample from over a decade's worth of tweets,
we track the rate of originally authored messages, retweets, and social amplification
for over 100 languages.

We describe distinct usage patterns of retweets for certain populations. 
For example, 
Thai, Korean, and Hindi have the highest contagion ratios,
while Japanese, Russian, Swedish, and Finish lie at the other end of the spectrum.
While there is a wide range of motives and practices associated with retweeting, 
our object of study is the simple differentiation of observed behavior 
between the act of replication of \textit{anything} 
and the act of \textit{de novo} generation
(i.e., between retweeted and what we will call organic messages). 

We acknowledge two important limitations from the start.
First, 
while it may be tempting to naively view ideas spreading as infectious diseases, 
the analogy falls well short of capturing the full gamut of social contagion mechanisms~\cite{goffman1964a,daley1965a,schelling1971a,granovetter1978a,dodds2004a,dodds2005a,centola2007b,ugander2012structural,cozzo2013contact,bessi2015science},
and a full understanding of social contagion remains to be established.
And second, 
while higher contagion ratios are in part due to active social amplification by users,
they may also, for example, 
reflect changes in Twitter's design of the retweet feature, 
changes in demographics, 
or changes in a population's 
general familiarity with social media.
Future work will shed light on the 
psychological and behavioral drivers for the use of retweets in each language
across geographical and societal markers,
including countries and communities. 

\subsection{Background and Motivation}
\label{sec:background}

Social contagion has been extensively studied across many disciplines including 
marketing~\cite{bass1969a,van2007new,trusov2009effects,iyengar2011opinion},
finance~\cite{kelly2000market,cipriani2008herd,hirshleifer2009thought,fenzl2012psychological},
sociology~\cite{hamilton1981models,bovasso1996network,fagan2007social},
and medicine~\cite{christakis2013social,papachristos2015tragic,pollack2017impact}.
Because it can be easier to access data on human social behavior from social media outlets 
than from other sources such as in-person or text-message conversations,
social contagion dynamics are often examined in the context of messages posted 
and subsequently re-posted on social media
platforms~\cite{bond201261,kramer2014experimental,ellison2014cultivating,ferrara2016rise}.
Indeed, 
the flow of information in the context of social contagion on digital media outlets, 
especially Twitter, has been widely studied over the last decade~\cite{lerman2010information,hodas2014simple},
with attention paid to the spreading of certain kinds of messages, such as 
rumours~\cite{borge2012absence,kwon2013prominent,ozturk2015combating,kaligotla2015agent,zubiaga2016analysing}, 
misinformation and ``fake news''~\cite{del2016spreading,spohr2017fake,shao2017spread,tornberg2018echo}.
Several models have also been proposed to predict the spread of information on Twitter~\cite{zaman2010predicting},
while other models have shown the differences in which various topics can propagate throughout social networks~\cite{romero2011differences,weng2012a}. 
Studies have also investigated the extent to which information spread on Twitter 
can have an echo chamber effect~\cite{colleoni2014echo,barbera2015tweeting,barbera2015birds}.

The body of research shows overwhelming evidence that retweeting is a key instrument of social contagion on 
Twitter~\cite{nagarajan2010qualitative,stieglitz2012political}.
One of the earliest analysis of Twitter by Kwak~et~al.~\cite{kwak2010twitter} 
suggests that a retweet can reach an average of a thousand users 
regardless of the social network of its original author, 
spreading its content instantly across different hubs of the full Twitter social network.
While seemingly simple, there are different styles and drivers of retweeting~\cite{boyd2010tweet}.
The practice of retweeting has become a convention on Twitter 
to spread information, especially for celebrities.
Researchers argue celebrities can act as hubs of social contagion 
by studying the flow of retweets across their focal networks~\cite{harrigan2012influentials}.
Recent work shows how retweets of officials can be either alarming or reassuring amid the COVID--19 pandemic~\cite{rao2020retweets,monsted2017evidence}. 
Statistical features of retweets reveal a strong association between links and hashtags in most retweeted messages~\cite{suh2010want}.
Retweeting is not only an act in which users can spread information,
but a mechanism for actors to become involved in a conversation without being active participants~\cite{boyd2010tweet}. 
The use of retweets empirically alters the visibility of information and how fast messages can spread on the platform~\cite{hodas2012visibility}.

Other studies have quantified language usage on social media~\cite{cha2010measuring,fitch2017empirical}, 
particularly on Twitter~\cite{bolhuis2010twitter,kim2014sociolinguistic}. 
While investigators have studied the use of retweets in the context of social contagion using network-based approaches~\cite{lerman2010information,romero2011differences,fabrega2013social,monsted2017evidence},
little research has been done regarding the statistical variability of retweets across the vast majority of languages. 
In this paper, 
by applying an updated language identification (LID) process 
to over a decade of Twitter messages, 
we explore a macroscopic description of social contagion
through the use of retweets across languages on Twitter. 
Our study addresses a unique property of social contagion on Twitter
by statistically quantifying the rate of retweets in each language. 
We show how the practice of retweeting 
varies across different languages and 
how retweeting naturally lends itself to micro-level discussions of social contagion
on Twitter, which can also be extended to other social media outlets with similar features.

\subsection{Overview}
\label{sec:overview}

We structure our paper as follows.
First, 
we discuss the state-of-the-art tools presently used for language detection 
of short and informal messages (e.g., tweets).
We then describe our dataset and processing pipeline to 
answer some key questions regarding social contagion through the use of retweets.
Based on our considerations, 
we deploy FastText-LID~\cite{joulin2016bag}
to identify and explore the evolution of 100+ languages 
in over $118$ billion messages collected via
Twitter's 10\% random sample (decahose) from 2009 to 2020~\cite{twitter_api}.

For messages posted after 2013, 
we also analyze language labels provided by Twitter's proprietary LID algorithm
and justify using FastText-LID as an alternative LID tool 
to overcome the challenge of missing language labels in the historical 
feed from Twitter (see also Hong~et~al.~\cite{hong2011language}).

We study the empirical dynamics of replication: 
The rate at which users choose to retweet instead of generating original content; 
and how that rate varies across languages temporally.
We quantify the ratio of retweets to new messages (contagion ratio) in each language. 
In most common languages on Twitter, 
we show that this ratio reveals a growing tendency to retweet.

Finally, 
we present a detailed comparison with the historical data feed in Appendix~\ref{sec:Decahose}.
We conclude with an analytical validation of our contagion ratios (Appendix~\ref{sec:validation}),
and the impact of tweet-length on language detection (Appendix~\ref{sec:tweet_len}).
We also provide an online appendix at:
\href{http://compstorylab.org/storywrangler/papers/tlid/}{http://compstorylab.org/storywrangler/papers/tlid}.

\section{Tweet Language Identification}
\label{sec:tlid}

Twitter is a well-structured streaming source of sociotechnical data,
allowing for the study of dynamical linguistics and cultural phenomena~\cite{zubiaga2015real,dewhurst2020shocklet}.
Of course, 
like many other social platforms, Twitter represents 
only a subsample of the publicly declared views, utterances, and interactions
of millions of individuals, organizations, 
and automated accounts (e.g., social bots) around the world~\cite{mellon2017twitter,ke2017scientistsonTwitter,mitchell2019reaction,wojcik2019sizing}. 
Researchers have nevertheless shown that Twitter's collective 
conversation mirrors the dynamics of local and global events~\cite{palen2016crisis} 
including
earthquakes~\cite{sakaki2010earthquake},
flu and influenza~\cite{lampos2010tracking,Culotta2010towards}, 
crowdsourcing and disaster relief~\cite{pickard2011time,gao2011harnessing},
major political affairs~\cite{steinert2015online},
and fame dynamics for political figures and celebrities~\cite{dodds2019fame}.
Moreover, analyses of social media data and digital text corpora over the last decade
have advanced natural language processing (NLP) research~\cite{ritter2011named,ritter12open,hirschberg2015advances} such as
language detection~\cite{lui2012langid,bergsma2012language,lui2014accuratelid,williams2017twitterlid},
sentiment analysis~\cite{dodds2011happiness,chu2012detecting,kharde2016sentiment,kryvasheyeu2016rapid,kursuncu2019predictive},
word embeddings~\cite{pennington2014glove,devlin2018bert,mikolov2018advances,grave2018learning},
and machine translation~\cite{papineni2002bleu,bahdanau2014neural,luong2015effective}.

LID is often referred to as a solved problem in NLP research~\cite{mcnamee2005language,hughes2006reconsidering,grothe2008comparative,lui2011cross,lui2014automatic}, especially for properly formatted documents, such as books, newspapers, and other long-form digital texts.
Language detection for tweets, however, is a challenging task due to the nature of the platform.
Every day, millions of text snippets are posted to Twitter and written in many languages along with
misspellings, catchphrases, memes, hashtags, and emojis, as well as images, gifs, and videos. 
Encoding many cultural phenomena semantically,
these features contribute to the unique aspects of language usage on Twitter 
that are distinct from studies of language on longer, edited corpora~\cite{michel176googlengrams}.  

A key challenge of LID on Twitter data is the absence of a large, public, 
annotated corpus of tweets covering most languages for training and evaluation of LID algorithms.
Although researchers have compiled a handful of manually labeled datasets of Twitter messages, 
the proposed datasets were notably small compared to the size of daily messages on Twitter 
and limited to a few common languages~\cite{bergsma2012language,lui2014accuratelid,williams2017twitterlid}.
They showed, however, that most off-the-shelf LID methods perform relatively well when tested on annotated tweets.

As of early 2013, Twitter introduced language predictions classified by their internal algorithm in the historical data feed~\cite{roomann2013lid}.
Since the LID algorithm used by Twitter is proprietary, 
we can only refer to a simple evaluation of their own model.\footnote{\url{https://blog.twitter.com/engineering/en_us/a/2015/evaluating-language-identification-performance.html}}
Our analysis of Twitter's language labels indicates 
Twitter appears to have tested several language detection methods, 
or perhaps different parameters, between 2013 and 2016.

Given access to additional information about the author of a tweet, 
the LID task would conceivably be much more accurate.  
For example, 
if the training data for prediction included any or all of the self-reported locations found in a user's `bio', the GPS coordinates of their most recent tweet, 
the language they prefer to read messages in, the language associated with individuals they follow or who follow them, 
and their collective tweet history, we expect the predictions would improve considerably. 
However, for the present investigation, 
we assume the only available predictors are found in the message itself.
Our goal is to use the state-of-the-art language detection tools 
to get consistent language labels for messages in our data set
to enable us to investigate broader questions about linguistic dynamics 
and the growth of retweets on the platform over time.

\subsection{Open-source Tools for LID}
\label{sec:lid_models}

Several studies have looked closely at language identification and detection for short-text~\cite{tromp2011graph,elfardy2012token,carter2013microblog,steinmetz2013,goldszmidt2013bootstrapping,nguyen2015audience,vilares2015,rijhwani2017}, 
particularly on Twitter where users are limited to a few characters per tweet 
(140 prior to the last few months of 2017, 280 thereafter~\cite{tweet_len}). 
Researchers have outlined common challenges specific to this platform~\cite{batrinca2015social,giachanou2016like}.

Most studies share a strong consensus that language identification of tweets is an exceptionally difficult task for several reasons. 
First, language classification models are usually trained  
over formal and large corpora, 
while most messages shared on Twitter 
are informal and composed of 140 characters or fewer~\cite{bergsma2012language,lui2014accuratelid}
(see Appendix~\ref{sec:tweet_len} for more details).
Second, 
the informal nature of the content is also a function of  
linguistic and cultural norms;
some languages are used differently over social media 
compared to the way they are normally used in books and formal documents. 
Third, 
users are not forced to choose a single language for each message; 
indeed messages are often posted with words from several languages found in a single tweet.
Therefore, the combination of short, informal, and multilingual posts on Twitter
makes language detection much more difficult compared to LID of formal documents~\cite{pla2017language}.
Finally, the lack of large collections of verified ground-truth across most languages 
is challenging for data scientists seeking to fine-tune language detection models using Twitter data~\cite{bergsma2012language,zubiaga2016tweetlid,blodgett2017dataset}.

Researchers have evaluated off-the-shelf LID tools on substantial subsets of Twitter data for a limited number of languages~\cite{bergsma2012language,lui2014accuratelid,blodgett2017dataset}. 
For example, Google's Compact Language Detector 
(versions 
CLD-1\footnote{\url{http://code.google.com/p/chromium-compact-language-detector/}} and 
CLD-2\footnote{\url{https://github.com/CLD2Owners/cld2}}) 
offer open-source implementations of the default LID tool in the Chrome browser to detect language used on web pages using a naive Bayes classifier. 
In 2012, Lui and Baldwin~\cite{lui2012langid} proposed a model called langid that uses an $n$-gram-based multinomial naive Bayes classifier. 
They evaluated langid and showed that it outperforms Google's CLD on multiple datasets. 
A majority-vote ensemble of LID models is also proposed by Lui~et~al.~\cite{lui2014accuratelid} 
that combines both Google's CLD and langid to improve classification accuracy for Twitter data. 
 
Although using a majority-vote ensemble of LID models may be the best option to maximize accuracy, 
there are a few critical trade-offs including speed and uncertainty. 
The first challenge of using an ensemble is weighing the votes of different models.
One can propose treating all models equally and taking the majority vote. 
This becomes evidently complicated in case of a tie, or when models are completely unclear on a given tweet.
Treating all models equally is an arguably flawed assumption given that not all models will have the same confidence 
in each prediction---if any is reported. 
Unfortunately, most LID models either decline to report a confidence score, 
or lack a clear and consistent way of measuring their confidence.
Finally, running multiple LID classifiers on every tweet is computationally expensive and time-consuming.

Recent advances in word embeddings powered by deep learning 
demonstrate some of the greatest breakthroughs across many NLP tasks including LID. 
Unlike previous methodologies,  
Devlin~et~al.~\cite{devlin2018bert} introduces a new language representation model called BERT. 
An additional output layer can be added to the pre-trained model 
to harvest the power of the distributed language representations,
which enables the model to carry out various NLP tasks such as LID. 

FastText~\cite{joulin2016bag} is a recently proposed approach 
for text classification that uses $n$-gram features similar to the model described by Mikolov~et~al.~\cite{mikolov2013efficient}. 
FastText employs various tricks~\cite{bojanowski2017enriching,mikolov2018advances,grave2018learning} 
in order to train a simple neural network using stochastic gradient descent 
and a linearly decaying learning rate for text classification. 
While FastText is a language model that can be used for various text mining tasks,
it requires an additional step of producing vector language representations to be used for LID. 
To accomplish that, we use an off-the-shelf language identification 
tool~\cite{fasttext} that uses the word embeddings produced by the model. 
The proposed tool uses a hierarchical softmax function~\cite{mikolov2013efficient,joulin2016bag} 
to efficiently compute the probability distribution over the predefined classes (i.e., languages). 
For convenience, we will refer to the off-the-shelf LID tool~\cite{fasttext} as FastText-LID throughout the paper. 
The authors show that FastText-LID is on par with deep learning models~\cite{xiao2016character,conneau2016very} in terms of accuracy and consistency,
yet orders of magnitude faster in terms of inference and training time~\cite{joulin2016bag}.
They also show that FastText-LID outperforms previously introduced LID tools such as 
langid.\footnote{\url{https://fasttext.cc/blog/2017/10/02/blog-post.html}}

\subsection{Processing Pipeline}
\label{sec:pipeline}

%\begin{algorithm}[H] 
%  \caption{Tweet Language Identification} 
%  \label{tlid}
%   \begin{algorithmic}[1]
%   \Require Tweet text
%   \Ensure Language iso-code
%   \State $text \gets \verb|Filter|(original\_text)$ \Comment{Twitter-specific content}
%   \State $\texttt{lang}, \texttt{conf} \gets FastText(text)$  \Comment{Language identification}
%   \If{$\texttt{conf} < .25$}
%        \State return \texttt{und}
%   \Else
%        \State return \texttt{lang}
%   \EndIf
%   \end{algorithmic} 
%\end{algorithm} 

While there are many tools to consider for LID,
it is important for us to ensure that the language classification process stays rather consistent 
to investigate our key question about the growth of retweets over time. 
In light of the technical challenges discussed in the previous section, 
we have confined this work to using FastText-LID~\cite{fasttext} due to its consistent 
and reliable performance in terms of inference time and accuracy.  

To avoid biasing our language classification process, 
we filter out Twitter-specific content prior to passing tweets through the FastText-LID model. 
This is a simple strategy originally proposed by Tromp~et~al.~\cite{tromp2011graph} to improve language classification~\cite{lui2014accuratelid,bergsma2013broadly}.
Specifically, we remove the prefix associated with retweets (``RT''), 
links (e.g., ``https://twitter.com''), 
hashtags (e.g., ``\#newyear''), 
handles (e.g., ``@username''), 
html codes (e.g., ``\&gt''), 
emojis, and any redundant whitespaces. 

Once we filter out all Twitter-specific content, we feed the remaining text through the FastText-LID neural network and select the predicted language with the highest confidence score as our ground-truth language label. 
If the confidence score of a given prediction is less than 25\%, we label that tweet as Undefined (\verb|und|).
Similarly, if no language classification is made by the Twitter-LID model, Twitter flags the language of the message as undefined~\cite{twitter_api_rules,bcp}.
We provide a list of all language labels assigned by FastText-LID compared to the ones served by Twitter-LID in Table~\ref{tab:iso-codes}.  

We subsequently extract day-scale time series and Zipf distributions
for uni-, bi-, and tri-grams 
and make them available through an analytical instrument entitled Storywrangler. 
Our tool is publicly available online at:
\href{https://storywrangling.org/}{https://storywrangling.org/}.
See Alshaabi~et~al.~\cite{alshaabi2020storywrangler} for technical details about our project.

\begin{figure}[tp!]
\includegraphics[width=\columnwidth]{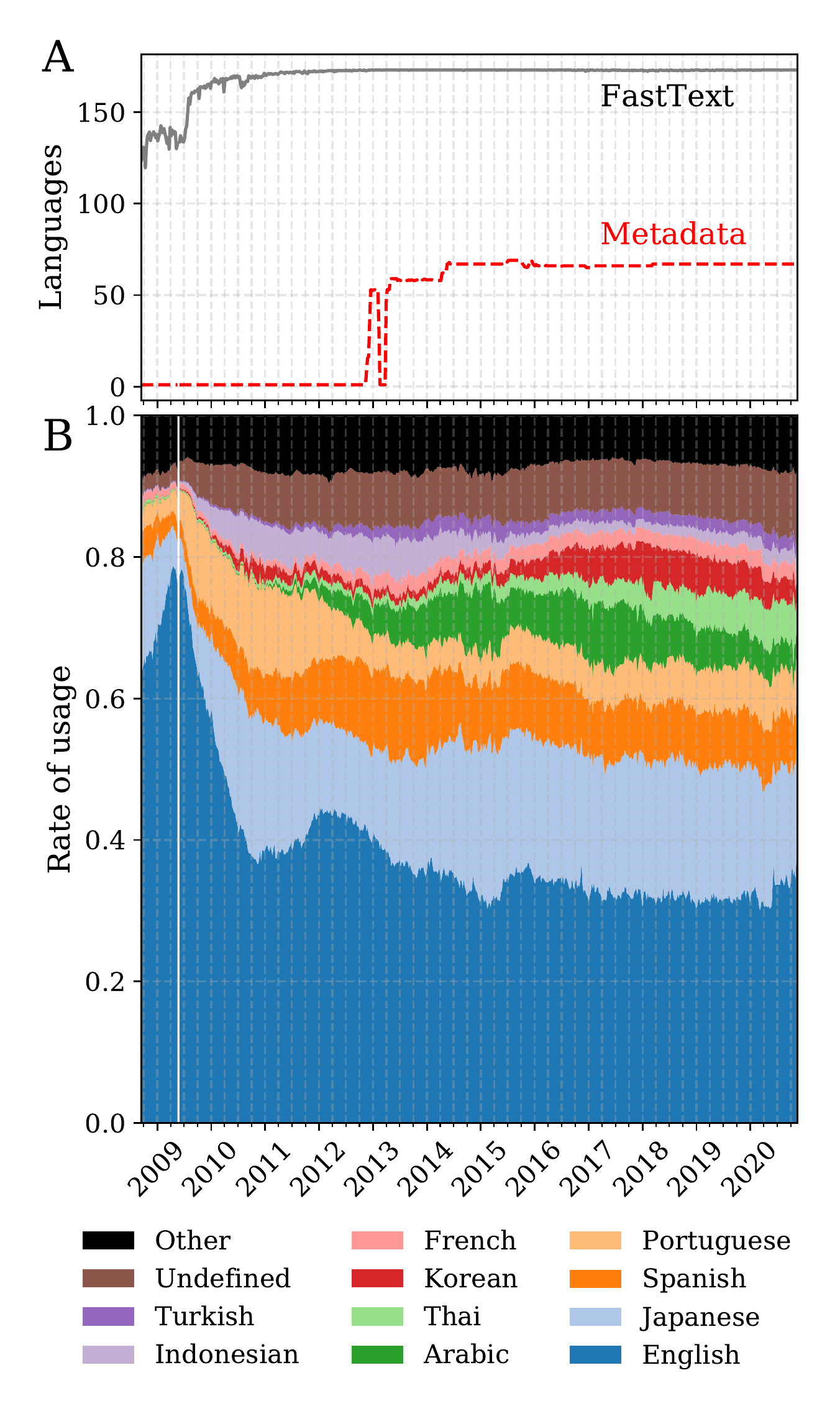}
\caption{
\textbf{Language time series for the Twitter historical feed and FastText-LID classified tweets.}
\textbf{A.} 
Number of languages reported by Twitter-LID (red) and classified by FastText-LID (black) 
since September 2008. 
Fluctuations in late 2012 and early 2013 for the Twitter language time series are indicative of inconsistent classifications.
\textbf{B.} 
Rate of usage by language using FastText-LID maintains consistent behavior throughout throughout that period.
The change in language distribution when Twitter was relatively immature can be readily seen---for instance, 
English accounted for an exceedingly high proportion of activity on the platform in 2009, 
owing to Twitter's inception in an English-speaking region.
}
\label{fig:language_ts}     
\end{figure}

\begin{figure*}[tp!]
\includegraphics[width=\textwidth]{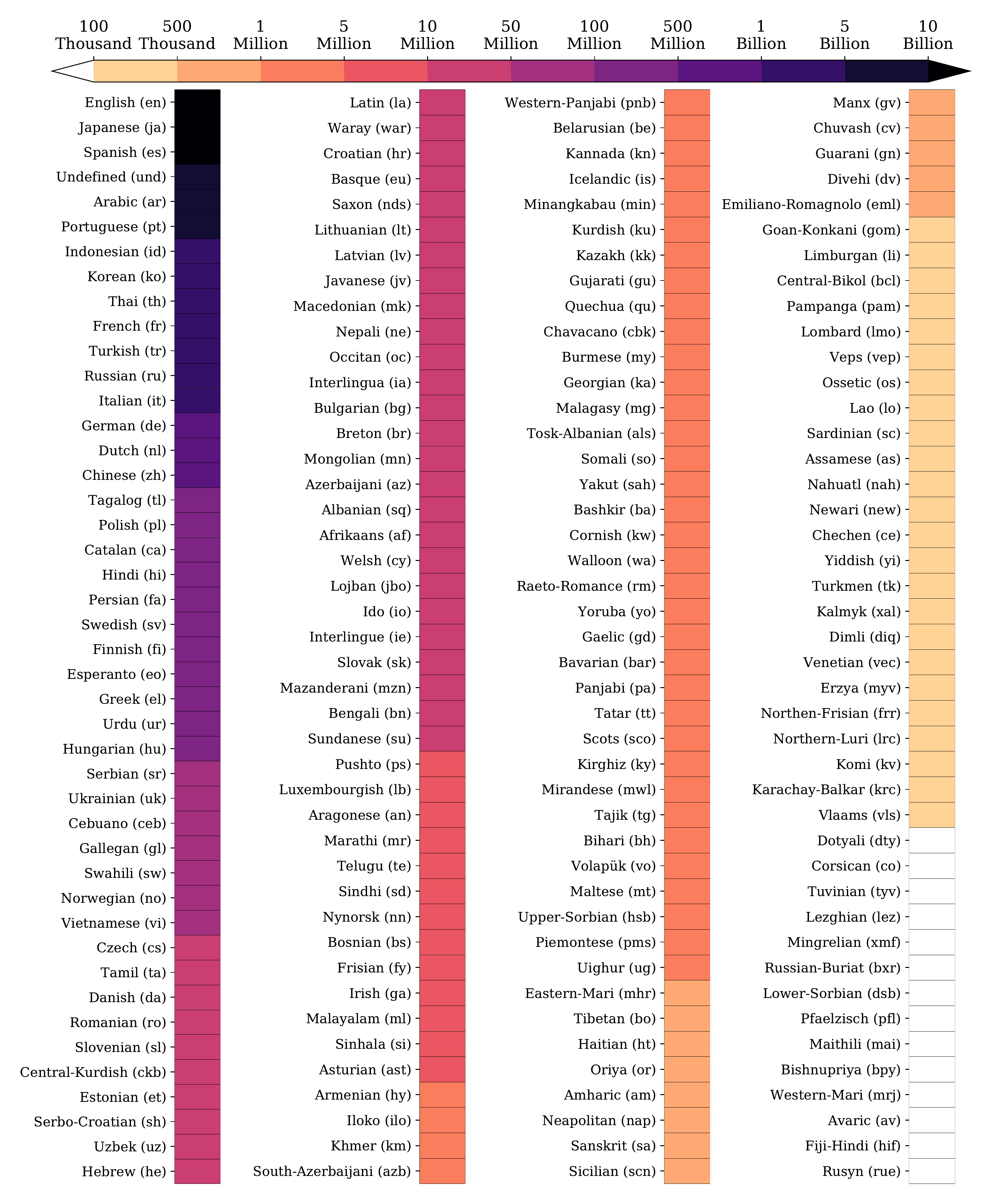}
\caption{
\textbf{Overall dataset statistics.}
Number of messages captured in our dataset 
as classified by the FastText-LID algorithm
between 2009-01-01 and 2019-12-31,
which sums up to a approximately 118 billion messages throughout that period 
(languages are sorted by popularity). 
This collection represents roughly 10\% of all messages ever posted.
}
\label{fig:stats}   
\end{figure*} 

\clearpage

\clearpage
\section{Results and Discussion}
\label{sec:results}

\subsection{Temporal and Empirical Statistics}
\label{sec:languages}
\begin{figure*}[tp!]
\includegraphics[width=\textwidth]{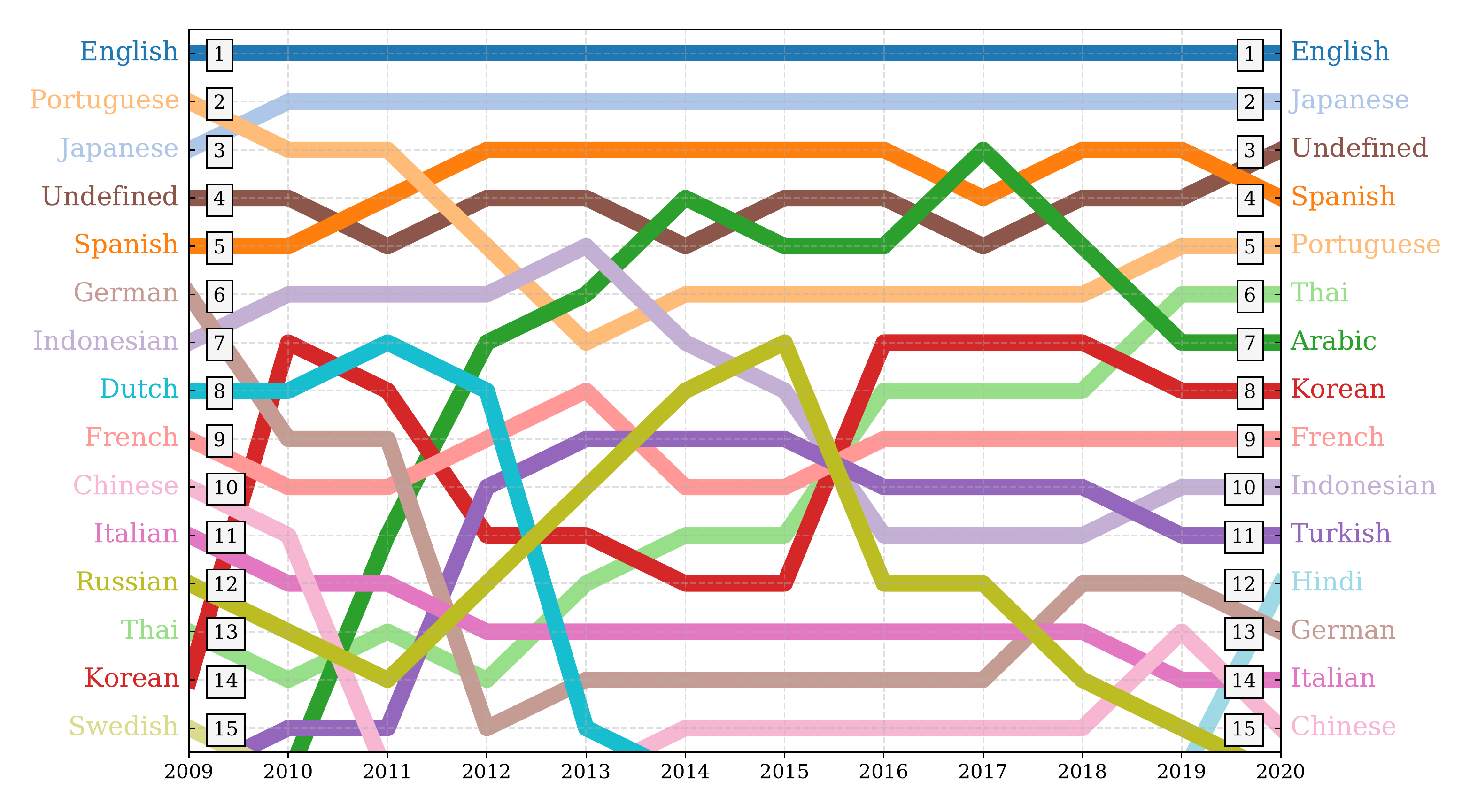}
\caption{ 
\textbf{Annual average rank of the most used languages on Twitter between 2009 and 2020.}
English and Japanese show the most consistent rank time series.
Spanish, and Portuguese are also relatively stable over time. 
Undefined---which covers a wide variety of content such as emojis, links, pictures, and other media---also has a consistent rank time series. 
The rise of languages on the platform correlates strongly with 
international events including Arab Spring and K-pop, as evident in both the Arabic and Korean time series, respectively. 
Russian, German, Indonesian, and Dutch moved down in rank. 
This shift is not necessarily due to a dramatic drop in the rate of usage of these languages, 
but is likely an artifact of increasing growth of other languages on Twitter such as Thai, Turkish, Arabic, Korean, etc.
}
\label{fig:rank_ts}     
\end{figure*}

We have collected a random 10\% sample of all public tweets posted on the Twitter platform starting January 1, 2009. 
Using the steps described in Sec.~\ref{sec:pipeline}, 
we have implemented a simple pipeline to preprocess messages and
obtain language labels using FastText-LID~\cite{fasttext}.
Our source code along with our documentation is publicly available online on a Gitlab repository.\footnote{\url{https://gitlab.com/compstorylab/storywrangler}} 
Here, we evaluate our results by comparing the language labels obtained by FastText-LID to those found in the metadata provided by Twitter’s internal LID algorithm(s).
Our initial analysis of the Decahose metadata indicated missing language labels until 2013, when Twitter began offering a language prediction (we offer an approach to detecting corrupted time series in Dodds~\etal~\cite{dodds2020c}).

We find that our classification of tweets using FastText-LID notably improves the consistency 
of language labels when compared to the labels served with the historical streaming feed. 
In Fig.~\ref{fig:language_ts}A, 
we display a weekly rolling average of the daily number of languages detected by each classifier over time. 
We see that Twitter's language detection has evolved over time.
The number of languages stabilized but continued to fluctuate in a manner that is not consistent, 
with uncommon languages having zero observations on some given days. 
By contrast, the FastText-LID time series of the number of languages 
shows some fluctuations in the earlier 
years---likely the result of the smaller and less diverse user base in the late 2000s---but 
stabilizes before Twitter introduces language labels. 
We note that the fluctuations in the time series during the early years of Twitter 
(before 2012) and the first week of 2017 
are primarily caused by unexpected service outages which resulted in missing data.

FastText-LID classifies up to 173 languages,
some of which are rare, 
thus the occasional dropout of a language seen in this time series is expected. 
On the other hand, Twitter-LID captures up to 73 languages, 
some of which are experimental and no longer available in recent years.
Nonetheless, Fig.~\ref{fig:language_ts}B shows that the overall rate of usage by language 
is not impaired by the missing data, and maintained consistent behavior throughout the last decade. 

We compute annual confusion matrices to examine the language labels classified by FastText-LID compared to those found in the historical data feed. 
Upon inspection of the computed confusion matrices, 
we find disagreement during the first few years of Twitter's introduction of the LID feature to the platform.
As anticipated, the predicted language for the majority of tweets harmonizes across both classifiers for recent years (see Fig.~\ref{fig:confmat}).
We notice some disagreement between the two classifiers 
on expected edge-cases such as Italian, Spanish, and Portuguese 
where the lexical similarity among these languages is very high~\cite{ringbom2007cross,borer2014parametric,samoilenko2016linguistic,jin2017can}. 
Overall, our examination of average language usage over time demonstrates that 
FastText-LID is on par with Twitter's estimation. 
We show the corresponding Zipf distribution of language usage for each classifier,
and highlight the normalized ratio difference between them
for the most used languages on the platform
in Figs.~\ref{fig:divergence}--\ref{fig:shifts}. 
We point the reader's attention to Appendix~\ref{sec:Decahose}
for further details of our comparison. 

\begin{figure*}[tp!] 
\includegraphics[width=\textwidth]{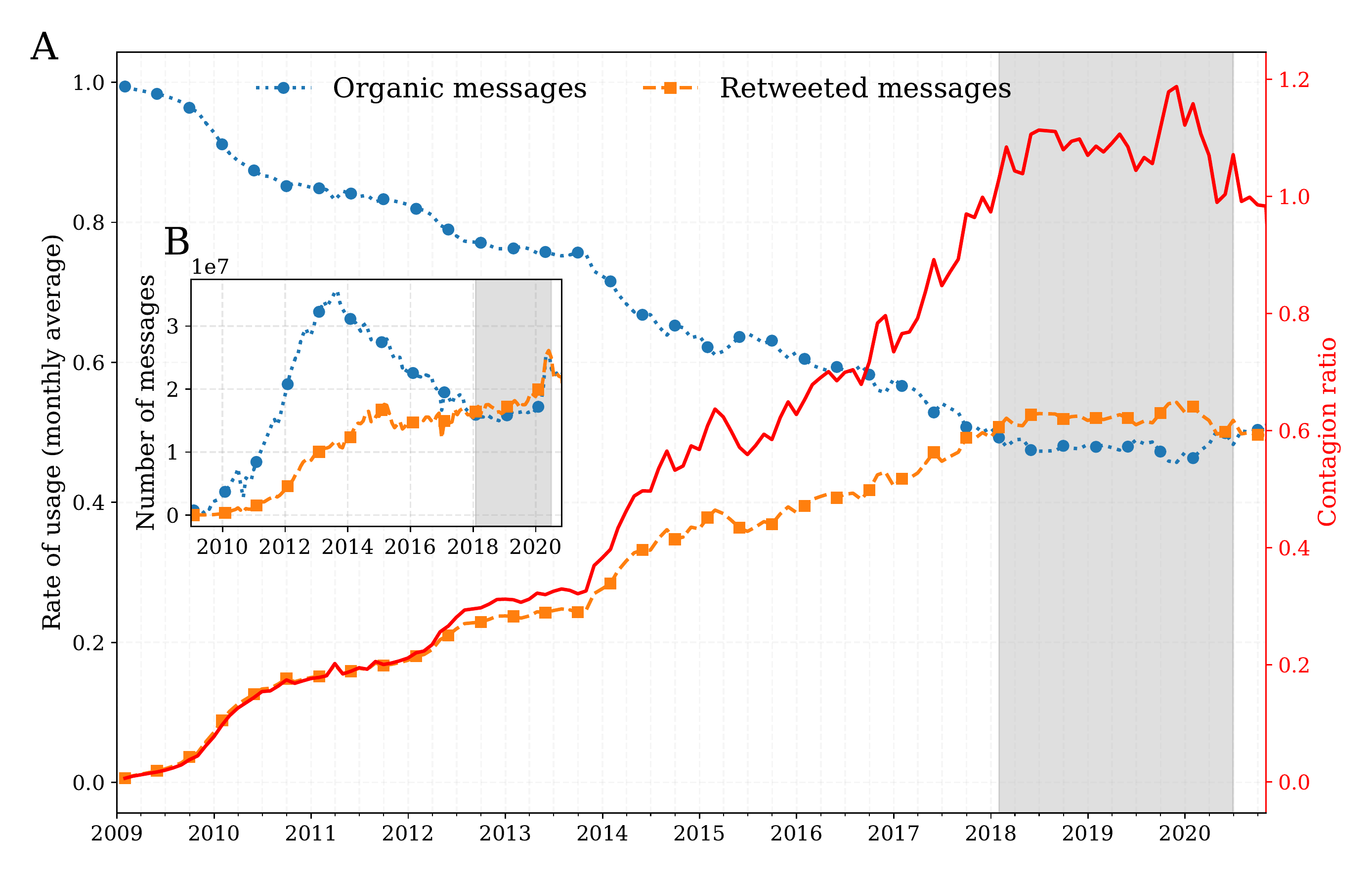}
\caption{
\textbf{Timeseries for organic messages, retweeted messages, and average contagion ratio for all languages.}
\textbf{A.} 
Monthly average rate of usage of 
organic messages ($p_{t,\ell}^{(\textnormal{OT})}$, blue), and
retweeted messages ($p_{t,\ell}^{(\textnormal{RT})}$, orange).
The solid red line highlights the steady rise of the contagion ratio $R_{\ell,t}$. 
\textbf{B.} 
Frequency of 
organic messages ($f^{\textnormal{(OT)}}_{\ell,t}$, blue), compared to
retweeted messages ($f^{\textnormal{(RT)}}_{\ell,t}$, orange).
The areas shaded in light grey starting in early 2018 highlights 
an interesting shift on the platform 
where the number of retweeted messages has exceeded the number of organic messages. 
An interactive version of the figure for all languages is available in an online appendix:
\href{http://compstorylab.org/storywrangler/papers/tlid/files/ratio\_timeseries.html}{http://compstorylab.org/storywrangler/papers/tlid/files/ratio\_timeseries.html}.
}
\label{fig:ratio_ts}    
\end{figure*}

Furthermore, 
we display a heatmap of the number of messages for each language as classified 
by FastText-LID in our data set (see Fig.~\ref{fig:stats}).
We have over 118 billion messages between 2009-01-01 and 2019-12-31 spanning 173 languages. 
English is the most used language on the platform with a little under 42 billion messages throughout the last decade.
Although the number of Japanese speakers is much smaller than
the number of English speakers around the globe, Japanese has approximately 21 billion messages.
Spanish---the third most prominent language on Twitter---is shy of 11 billion messages. 
Arabic and Portuguese rank next with about 7 billion messages for each.
We note that the top 10 languages comprise 85\% of the daily volume of messages posted on the platform.

In Fig.~\ref{fig:rank_ts}, we show the flow of annual rank dynamics of the 15 most used languages on Twitter between 2009 and 2020.
For ease of description, we will refer to Undefined as a language class. 
The top 5 most common languages on Twitter (English, Japanese, Spanish, Undefined, and Portuguese) are consistent, indicating a steady rate of usage of these languages on the platform. 
The language rankings correspond with worldwide events such as the
Arab Spring~\cite{howard2013democracy,wolfsfeld2013social,dewey2012impact,cottle2011media},
K-pop, and political events~\cite{dodds2019fame}.
``Undefined'' is especially interesting as it covers a wide range of content such as emojis, memes, and other media shared on Twitter but can't necessarily be associated with a given language. 
Russian, however, starts to grow on the platform after 2011 until it peaks with a rank of 7 in 2015, then drops down to rank 15 as of the end of 2019.  
Other languages such as German, Indonesian, and Dutch show a similar trend down in ranking.
This shift is not necessarily caused by a drop in the rate of usage of these languages, 
but it is rather an artifact prompted by the growth of other languages on Twitter.

\subsection{Quantifying Twitter's Social Contagion: Separating Organic and Retweeted Messages}
\label{sec:echo_chamber}

\begin{figure*}[tp!]
\centerline{\includegraphics[width=1.1\textwidth]{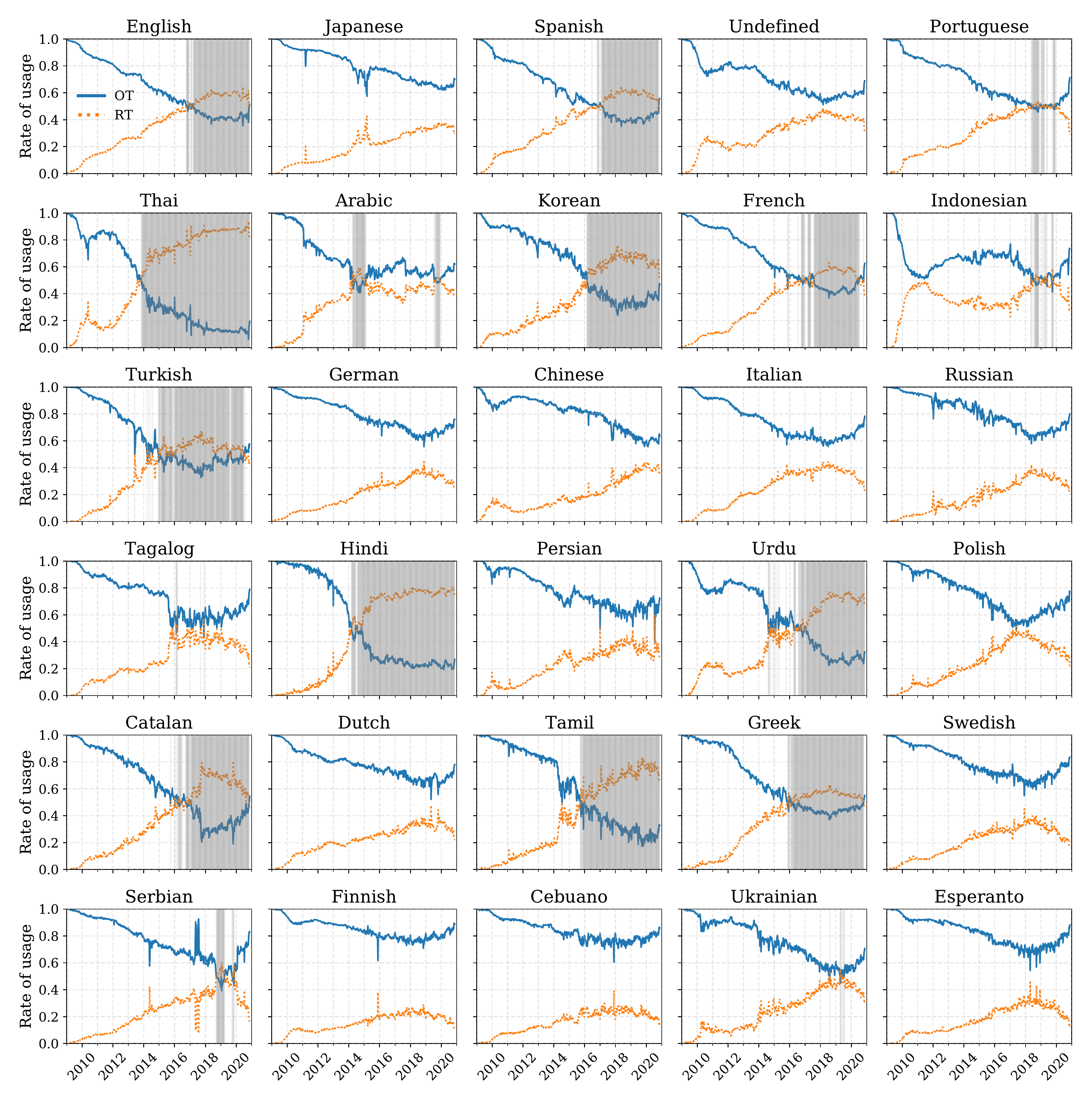}}
\caption{
\textbf{Weekly rate of usage of the top 30 languages (sorted by popularity).}
For each language, we show a weekly average rate of usage for 
organic messages ($p_{t,\ell}^{(\textnormal{OT})}$, blue)
compared to retweeted messages ($p_{t,\ell}^{(\textnormal{RT})}$, orange). 
The areas highlighted in light shades of gray represent 
weeks where the rate of retweeted messages is higher than the rate of organic messages.
An interactive version featuring all languages is available in an online appendix: 
\href{http://compstorylab.org/storywrangler/papers/tlid/files/retweets\_timeseries.html}{http://compstorylab.org/storywrangler/papers/tlid/files/retweets\_timeseries.html}.
} 
\label{fig:ratio_lang_ts} 
\end{figure*}

We take a closer look at the flow of information among different languages on the platform, 
specifically the use of the ``retweet'' feature as a way of spreading information. 
Encoding a behavioral feature initially invented by users,
Twitter formalized the retweet feature in November 2009~\cite{stone2009RT}. 
Changes in platform design and the increasing popularity of mobile apps promoted the RT as a mechanism for spreading.
In April 2015, Twitter introduced the ability to comment on a retweet message or 
``Quote Tweet''(QT)~\cite{Shu2015QT} a message,
distinct from a message reply~\cite{stone2007RP}. 

To quantify the rate of usage of each language with respect to these different means by which people communicate on the platform, we categorize messages on Twitter into two types:
``Organic Tweets'' (OT), and ``Retweets'' (RT).
The former category (OT) encompasses original messages that are explicitly authored by users,
while the latter category (RT) captures messages that are shared (i.e.~amplified) by users.
We break each quote tweet into two separate messages: a comment and a retweet.  
We exclude retweets while including all added text (comments) found in quote tweets for the OT category. 

For each day $t$ and for each language $\ell$, 
we calculate the raw frequency (count) of organic messages $f^{\textnormal{(OT)}}_{\ell,t}$,
and retweets $f^{\textnormal{(RT)}}_{\ell,t}$.
We further determine the frequency of all tweets (AT) such that:
$f^{\textnormal{(AT)}}_{\ell,t} = f^{\textnormal{(OT)}}_{\ell,t} + f^{\textnormal{(RT)}}_{\ell,t}$.
The corresponding rate of usages (normalized frequencies) for these two categories are then:
\begin{equation}\label{eq:f_ot_rt}
  p_{t,\ell}^{(\textnormal{OT})}
  =
  \frac{f_{t,\ell}^{(\textnormal{OT})}}{f_{t,\ell}^{(\textnormal{AT})}},
  \
  \textnormal{and}
  \
  p_{t,\ell}^{(\textnormal{RT})}
  =
  \frac{f_{t,\ell}^{(\textnormal{RT})}}{f_{t,\ell}^{(\textnormal{AT})}}.
\end{equation}

\subsection{Measuring Social and Linguistic Wildfire through the Growth of Retweets}
\label{sec:ratio}  
\begin{figure}[tp!] 
\includegraphics[width=\columnwidth]{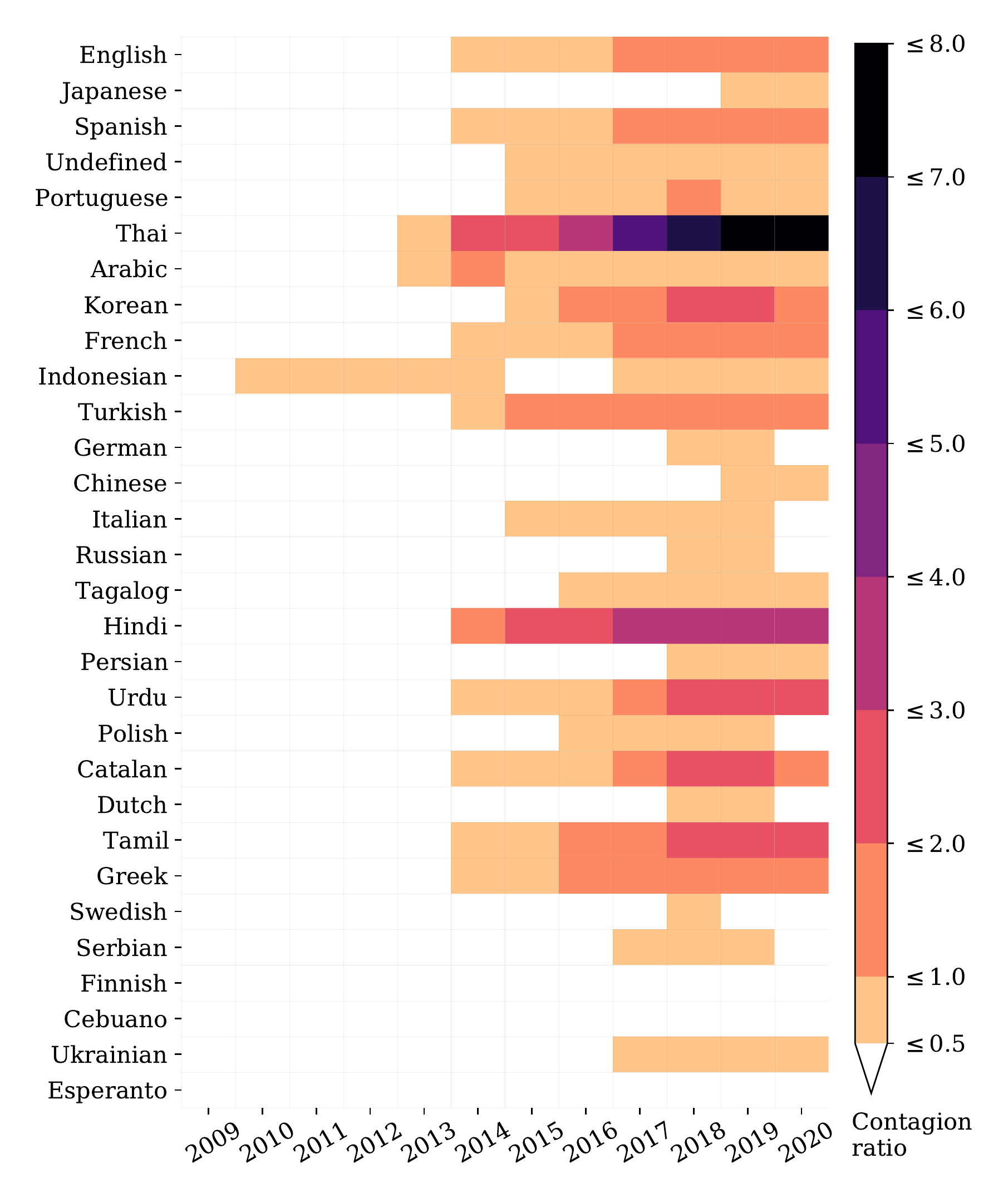}
\caption{\textbf{Timelapse of contagion ratios.}
The average ratio is plotted against year for the top 30 ranked languages of 2019.
Colored cells indicate a ratio higher than 0.5 whereas ratios below 0.5 are colored in white.
Table~\ref{tab:top_ratios} shows the top 10 languages with the highest average contagion ratio per year,
while Table~\ref{tab:bottom_ratios} shows the bottom 10 languages with the lowest average contagion ratio per year.
} 
\label{fig:ratio_heatmap}     
\end{figure}

To further investigate the growth of retweets, 
we use the ratio of retweeted messages to organic messages 
as an intuitive and interpretable analytical measure 
to track this social amplification phenomenon.
We define the `contagion ratio' as: 
\begin{align}\label{eq:contagion_ratio}
    R_{\ell,t} = f^{\textnormal{(RT)}}_{\ell,t} \bigg/ f^{\textnormal{(OT)}}_{\ell,t}.
\end{align}

In 2018, 
the contagion ratio exceeded 1, 
indicating a higher number of retweeted messages 
than organic messages
(Fig.~\ref{fig:ratio_ts}).
The overall count for organic messages peaked in the last quarter of 2013, 
after which it declined slowly as the number of retweeted messages climbed to approximately 1.2 retweeted messages 
for every organic message at the end of 2019. 
Thereafter, the contagion ratio declined through 2020
with the exception of a surge of retweets in the summer amid the 
nationwide protests sparked by the murder of George Floyd.\footnote{\url{https://www.nytimes.com/2020/05/31/us/george-floyd-investigation.html}}   

In 2020, Twitter's developers redesigned their retweet mechanism,
purposefully prompting users to write their own commentary using the
Quote Tweet~\cite{retweet_new_layout},
along with several new policies to counter synthetic and manipulated media~\cite{new_rules,misleading_information,new_policies}.
While the long upward trend of the contagion ratio is in part due to increasingly active social amplification by users,
the recent trend demonstrates how social amplification on Twitter is highly susceptible to systematic changes in the platform design.  
Twitter has also introduced several features 
throughout the last decade, such as 
tweet ranking,
and extended tweet length
that may intrinsically influence how users receive and share information in their social networks.\footnote{\url{https://help.twitter.com/en/using-twitter/twitter-conversations}} 
We investigate the robustness of our findings regarding contagion ratios 
in light of some of these changes in 
Appendix~\ref{sec:validation} and Appendix~\ref{sec:tweet_len}.
Future work will shed light on various aspects of social amplification on Twitter 
with respect to the evolution of the platform design,
and behavioral drivers for the use of retweets in each language across communities. 

Finally, 
we show weekly aggregation of the rate of usage of 
the top 30 ranked languages of 2019 in Fig.~\ref{fig:ratio_lang_ts}. 
The time series demonstrate a recent sociolinguistic shift: 
Several languages including English, Spanish, Thai, Korean, and French have transitioned to having a higher rate of retweeted messages than organic messages. 
Thai appears to be the first language to have made this transition in late 2013.
In Fig.~\ref{fig:ratio_heatmap}, 
we show a heatmap of the average contagion ratio for 
the top 30 most used languages on Twitter per year. 
With the exception of Indonesian,
which showed a small bump between 2010 and 2013, 
most other languages began adopting a higher ratio of retweeted content in 2014.
Thai has the highest number of retweeted messages, 
with an average of 7 retweeted messages for every organic message. 
Other languages, for example, Hindi, Korean, Urdu, Catalan, and Tamil
average between 2 to 4 retweeted messages for every organic message. 
On the other hand, 
Japanese---the second most used language on the platform---does not exhibit this trend.
Similarly, German, Italian, and Russian
maintain higher rates of organic tweets. 
The trend of increasing preference for retweeted messages,
though not universal, is evident among most languages on Twitter.
We highlight the top 10 languages with the highest and lowest average contagion ratio per year 
in Table~\ref{tab:top_ratios} and Table~\ref{tab:bottom_ratios}, respectively.

\section{Concluding Remarks}
\label{sec:conclusion}
Understanding how stories spread through and persist within populations
has always been central to understanding social phenomena.
In a time when information can flow instantly and freely 
online, the study of social contagion has only become more important.

In the sphere of Twitter, the practice of retweeting is complicated 
from a social and psychological point of view.
There is a diverse set of reasons for participants to retweet.
For example, scientists and academics can use this elementary feature to share their findings and discoveries with their colleagues. 
Celebrities and political actors can benefit from other people retweeting their stories for self-promotion. 
Attackers can also take advantage of this natural feature of social contagion to pursue malicious intents, 
deploy social bots, and spread fake news. 

In this paper, 
we have analyzed over a hundred billion messages posted on Twitter throughout the last decade.
We presented an alternative approach for obtaining language labels using FastText-LID in order to overcome the challenge of missing labels in the Decahose dataset, 
obtaining consistent language labels for 100+ languages.
We acknowledge that shortcomings of language detection for short and informal text 
(e.g., tweets) are well known in the NLP literature. 
Using FastText-LID is not necessarily the best approach for language identification. 
Our analysis may be subject to implicit measurement biases and errors 
introduced by word embeddings used to train the 
language detection tool using FastText~\cite{joulin2016bag}. 
We emphasize that we have not intended to reinvent or improve FastText-LID in this work; 
we have used FastText-LID only as a (well-established and tested) tool 
to enable the study of social contagion dynamics on Twitter.
Nevertheless, 
we have presented some further analysis of FastText-LID compared to Twitter-LID in Appendix~\ref{sec:Decahose}.
Future work will undoubtedly continue
to improve language detection for short text, 
particularly for social media platforms.

Our results comparing language usage over time suggest a systematic shift on Twitter.
We found a recent tendency among most languages to increasingly retweet (spread information) rather than generate new content.
Understanding the general rise of retweeted messages requires further investigation. 
Possible partial causes might lie in changes in the design of the platform, 
increases in bot activity,
a fundamental shift in human information processing as social media becomes more familiar to users,
and changes in the demographics of users (e.g., younger users joining the platform). 

The metrics we have used to compute our contagion ratios are simple but rather limited. 
We primarily focused on tracking the rate of organic tweets and retweets
to quantify social amplification of messages on the platform.
While our approach of measuring the statistical properties of contagion ratios is important,
it falls short of capturing how retweets propagate throughout the social networks of users.
Future work may deploy a network-based approach to investigate the flow of retweets among users and followers. 
Whether or not the information is differentially propagated across languages, social groups, economic strata, or geographical regions is an important question for future research, and beyond the scope of our present work.

Geolocation information for Twitter is also limited, and 
here we have only examined contagion ratios at the language level. 
Language, transcending borders as it does, can nevertheless 
be used differently across communities. 
For instance, characterizing the temporal dynamics of contagion ratios for 
English,
which is used all around the globe, is very different from doing so for 
Thai---a language that is used within a geographically well-defined population. 
Different social and geographical communities have cultures of communication which will need to be explored in future work.

In particular, it is important to study the relationship between social contagion dynamics, geographical region, and language. It might be the case that contagion dynamics are more homogeneous across geographic regions even when each geographical region displays high language diversity, or \textit{vice versa}. 
However, in order to conduct this line of research, it is necessary to have accurate geotagging of tweets, 
which is currently not the case except for a very small subsample~\cite{twitter_geotags}.
Future research could focus on implementing accurate geotagging algorithms that assign tweets a probabilistic geographical location based on their text and user metadata, while fully respecting privacy through judicious use of masking algorithms.

\acknowledgments  
The authors are grateful for the computing resources 
provided by the Vermont Advanced Computing Core 
and financial support from the Massachusetts Mutual Life Insurance Company 
and Google Open Source under the Open-Source Complex Ecosystems And Networks (OCEAN) project.
Computations were performed on the Vermont Advanced Computing Core 
supported in part by NSF award No. OAC-1827314.
We thank Colin Van Oort and Anne Marie Stupinski for their comments on the manuscript.  

\bibliography{references}
%%%%%%%%%%%%%%%%%%%%%%%%%%%%%%%%%%%%%%%%%%%%%%%%%%%%%%%%%%%

\clearpage
\onecolumngrid
%% supplementary

%% following records the starting page number of the supplementary
%% section in a file called startsupp.txt
%% enables script-based breaking of manuscript and supplementary
\newwrite\tempfile
\immediate\openout\tempfile=startsupp.txt
\immediate\write\tempfile{\thepage}
\immediate\closeout\tempfile

\renewcommand{\thepage}{S\arabic{page}}
\renewcommand{\thesection}{S\arabic{section}}
\renewcommand{\thefigure}{S\arabic{figure}}
\renewcommand{\thetable}{S\arabic{table}}
\setcounter{page}{1}
\setcounter{section}{0}
\setcounter{figure}{0}
\setcounter{table}{0}
\setcounter{footnote}{0}

\section{Comparison with the historical feed} 
\label{sec:Decahose}

\begin{table*}[tp!]
\centering
\caption{Language codes for both FastText-LID and Twitter-LID tools \label{tab:iso-codes}}
\begin{scriptsize}
\begin{tabular}{l|cc||l|cc||l|cc}
\hline\hline
\textbf{Language} & \textbf{FastText} & \textbf{Twitter} & 
\textbf{Language} & \textbf{FastText} & \textbf{Twitter} & 
\textbf{Language} & \textbf{FastText} & \textbf{Twitter} \\
\hline
\textbf{Afrikaans} & af & - & \textbf{Haitian} & ht & ht & \textbf{Pfaelzisch} & pfl & - \\
\textbf{Albanian} & sq & - & \textbf{Hebrew} & he & he & \textbf{Piemontese} & pms & - \\
\textbf{Amharic} & am & am & \textbf{Hindi} & hi & hi & \textbf{Polish} & pl & pl \\
\textbf{Arabic} & ar & ar & \textbf{Hungarian} & hu & hu & \textbf{Portuguese} & pt & pt \\
\textbf{Aragonese} & an & - & \textbf{Icelandic} & is & is & \textbf{Pushto} & ps & ps \\
\textbf{Armenian} & hy & hy & \textbf{Ido} & io & - & \textbf{Quechua} & qu & - \\
\textbf{Assamese} & as & - & \textbf{Iloko} & ilo & - & \textbf{Raeto-Romance} & rm & - \\
\textbf{Asturian} & ast & - & \textbf{Indonesian} & id & in & \textbf{Romanian} & ro & ro \\
\textbf{Avaric} & av & - & \textbf{Inuktitut} & - & iu & \textbf{Russian-Buriat} & bxr & - \\
\textbf{Azerbaijani} & az & - & \textbf{Interlingua} & ia & - & \textbf{Russian} & ru & ru \\
\textbf{Bashkir} & ba & - & \textbf{Interlingue} & ie & - & \textbf{Rusyn} & rue & - \\
\textbf{Basque} & eu & eu & \textbf{Irish} & ga & - & \textbf{Sanskrit} & sa & - \\
\textbf{Bavarian} & bar & - & \textbf{Italian} & it & it & \textbf{Sardinian} & sc & - \\
\textbf{Belarusian} & be & - & \textbf{Japanese} & ja & ja & \textbf{Saxon} & nds & - \\
\textbf{Bengali} & bn & bn & \textbf{Javanese} & jv & - & \textbf{Scots} & sco & - \\
\textbf{Bihari} & bh & - & \textbf{Kalmyk} & xal & - & \textbf{Serbian} & sr & sr \\
\textbf{Bishnupriya} & bpy & - & \textbf{Kannada} & kn & kn & \textbf{Serbo-Croatian} & sh & - \\
\textbf{Bosnian} & bs & bs & \textbf{Karachay-Balkar} & krc & - & \textbf{Sicilian} & scn & - \\
\textbf{Breton} & br & - & \textbf{Kazakh} & kk & - & \textbf{Sindhi} & sd & sd \\
\textbf{Bulgarian} & bg & bg & \textbf{Khmer} & km & km & \textbf{Sinhala} & si & si \\
\textbf{Burmese} & my & my & \textbf{Kirghiz} & ky & - & \textbf{Slovak} & sk & - \\
\textbf{Catalan} & ca & ca & \textbf{Komi} & kv & - & \textbf{Slovenian} & sl & sl \\ 
\textbf{Cebuano} & ceb & - & \textbf{Korean} & ko & ko & \textbf{Somali} & so & - \\
\textbf{Cherokee} & - & chr & \textbf{Kurdish} & ku & - & \textbf{Shona} & - & sn \\
\textbf{Central-Bikol} & bcl & - & \textbf{Lao} & lo & lo & \textbf{South-Azerbaijani} & azb & - \\
\textbf{Central-Kurdish} & ckb & ckb & \textbf{Latin} & la & - & \textbf{Spanish} & es & es \\
\textbf{Chavacano} & cbk & - & \textbf{Latvian} & lv & lv & \textbf{Sundanese} & su & - \\
\textbf{Chechen} & ce & - & \textbf{Lezghian} & lez & - & \textbf{Swahili} & sw & - \\
\textbf{Chinese-Simplified} & - & zh-cn & \textbf{Limburgan} & li & - & \textbf{Swedish} & sv & sv \\
\textbf{Chinese-Traditional} & - & zh-tw & \textbf{Lithuanian} & lt & lt & \textbf{Tagalog} & tl & tl \\
\textbf{Chinese} & zh & zh & \textbf{Lojban} & jbo & - & \textbf{Tajik} & tg & - \\
\textbf{Chuvash} & cv & - & \textbf{Lombard} & lmo & - & \textbf{Tamil} & ta & ta \\
\textbf{Cornish} & kw & - & \textbf{Lower-Sorbian} & dsb & - & \textbf{Tatar} & tt & - \\
\textbf{Corsican} & co & - & \textbf{Luxembourgish} & lb & - & \textbf{Telugu} & te & te \\
\textbf{Croatian} & hr & - & \textbf{Macedonian} & mk & - & \textbf{Thai} & th & th \\
\textbf{Czech} & cs & cs & \textbf{Maithili} & mai & - & \textbf{Tibetan} & bo & bo \\
\textbf{Danish} & da & da & \textbf{Malagasy} & mg & - & \textbf{Tosk-Albanian} & als & - \\
\textbf{Dimli} & diq & - & \textbf{Malayalam} & ml & ml & \textbf{Turkish} & tr & tr \\
\textbf{Divehi} & dv & dv & \textbf{Malay} & ms & msa & \textbf{Turkmen} & tk & - \\
\textbf{Dotyali} & dty & - & \textbf{Maltese} & mt & - & \textbf{Tuvinian} & tyv & - \\
\textbf{Dutch} & nl & nl & \textbf{Manx} & gv & - & \textbf{Uighur} & ug & ug \\
\textbf{Eastern-Mari} & mhr & - & \textbf{Marathi} & mr & mr & \textbf{Ukrainian} & uk & uk \\
\textbf{Egyptian-Arabic} & arz & - & \textbf{Mazanderani} & mzn & - & \textbf{Upper-Sorbian} & hsb & - \\
\textbf{Emiliano-Romagnolo} & eml & - & \textbf{Minangkabau} & min & - & \textbf{Urdu} & ur & ur \\
\textbf{English} & en & en & \textbf{Mingrelian} & xmf & - & \textbf{Uzbek} & uz & - \\
\textbf{Erzya} & myv & - & \textbf{Mirandese} & mwl & - & \textbf{Venetian} & vec & - \\
\textbf{Esperanto} & eo & - & \textbf{Mongolian} & mn & - & \textbf{Veps} & vep & - \\
\textbf{Estonian} & et & et & \textbf{Nahuatl} & nah & - & \textbf{Vietnamese} & vi & vi \\
\textbf{Fiji-Hindi} & hif & - & \textbf{Neapolitan} & nap & - & \textbf{Vlaams} & vls & - \\
\textbf{Filipino} & - & fil & \textbf{Nepali} & ne & ne & \textbf{Volapük} & vo & - \\
\textbf{Finnish} & fi & fi & \textbf{Newari} & new & - & \textbf{Walloon} & wa & - \\
\textbf{French} & fr & fr & \textbf{Northen-Frisian} & frr & - & \textbf{Waray} & war & - \\
\textbf{Frisian} & fy & - & \textbf{Northern-Luri} & lrc & - & \textbf{Welsh} & cy & cy \\
\textbf{Gaelic} & gd & - & \textbf{Norwegian} & no & no & \textbf{Western-Mari} & mrj & - \\
\textbf{Gallegan} & gl & - & \textbf{Nynorsk} & nn & - & \textbf{Western-Panjabi} & pnb & - \\
\textbf{Georgian} & ka & ka & \textbf{Occitan} & oc & - & \textbf{Wu-Chinese} & wuu & - \\
\textbf{German} & de & de & \textbf{Oriya} & or & or & \textbf{Yakut} & sah & - \\
\textbf{Goan-Konkani} & gom & - & \textbf{Ossetic} & os & - & \textbf{Yiddish} & yi & - \\
\textbf{Greek} & el & el & \textbf{Pampanga} & pam & - & \textbf{Yoruba} & yo & - \\
\textbf{Guarani} & gn & - & \textbf{Panjabi} & pa & pa & \textbf{Yue-Chinese} & yue & - \\
\textbf{Gujarati} & gu & gu & \textbf{Persian} & fa & fa & \textbf{Undefined} & und & und \\
\hline\hline
\end{tabular}
\end{scriptsize}
\end{table*}

\begin{figure*}[th!] 
\includegraphics[width=\textwidth]{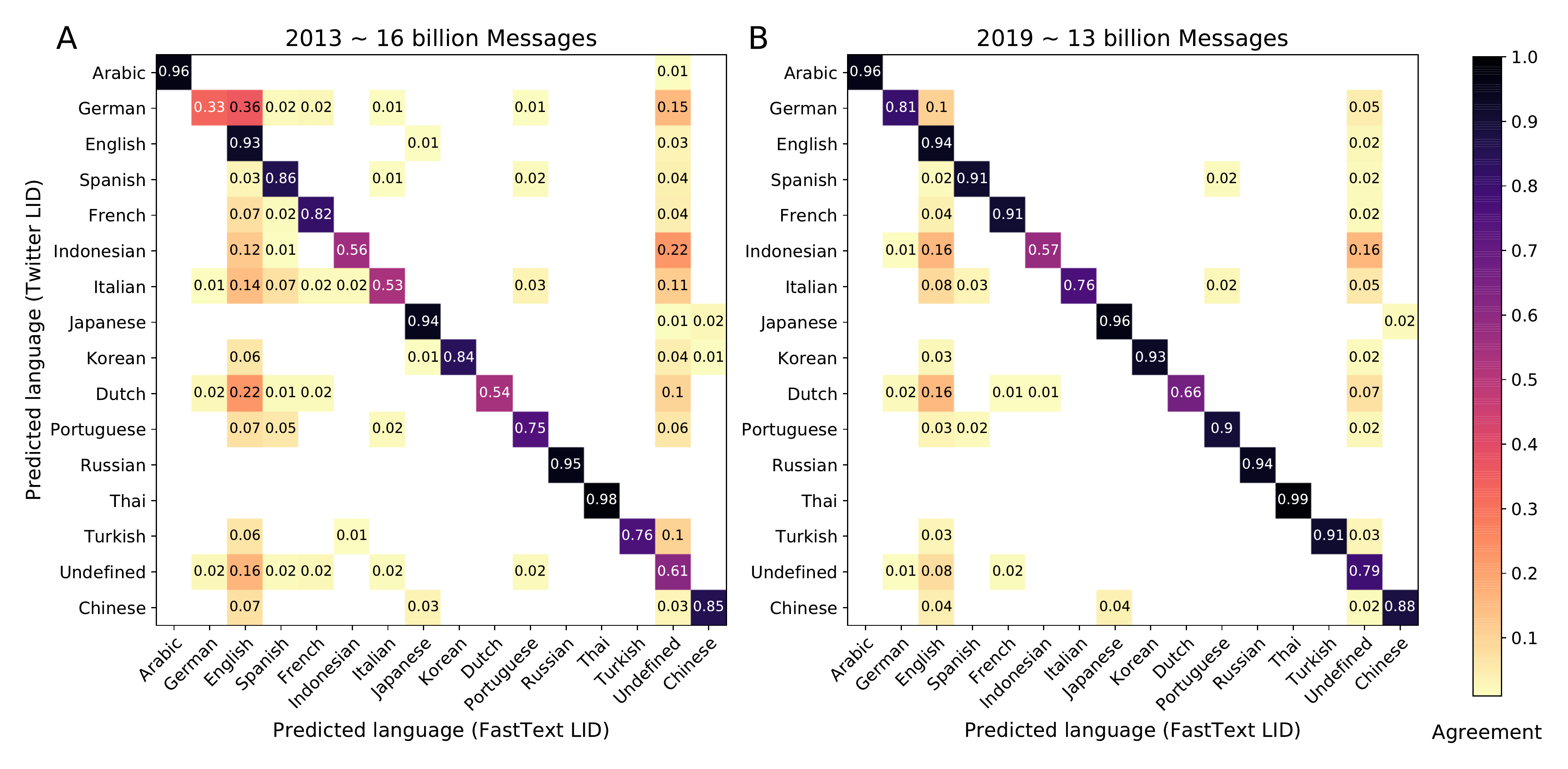} 
\caption{
\textbf{Language identification confusion matrices.}
We show a subset of the full confusion matrix for top-15 languages on Twitter.
\textbf{A.} 
Confusion matrix for tweets authored in 2013. 
The matrix indicates substantial disagreement between the two classifiers during 2013, 
the first year of Twitter's efforts to provide language labels.
\textbf{B.} 
For the year 2019, 
both classifiers agree on the majority of tweets as indicated by the dark diagonal line in the matrix.
Minor disagreement between the two classifiers is evident for particular languages, 
including German, Italian, and Undefined, 
and there is major disagreement for Indonesian and Dutch.
Cells with values below (.01) are colored in white 
to indicate very minor disagreement between the two classifiers. 
}   
\label{fig:confmat}   
\end{figure*}

We have collected all language labels served in the historical data feed, 
along with the predicted language label classified by FastText-LID 
for every individual tweet in our dataset. 
We provide a list of all language labels assigned by FastText-LID compared to the ones served by Twitter-LID in Table~\ref{tab:iso-codes}. 
To evaluate the agreement between the two classifiers, 
we computed annual confusion matrices starting from 2013 to the end of 2019.
In Fig.~\ref{fig:confmat}, 
we show confusion matrices for the 15 most dominate languages on Twitter 
for all tweets authored in 2013 (Fig.~\ref{fig:confmat}A) 
and 2019 (Fig.~\ref{fig:confmat}B).

\begin{figure*}[tp!] 
\includegraphics[width=\textwidth]{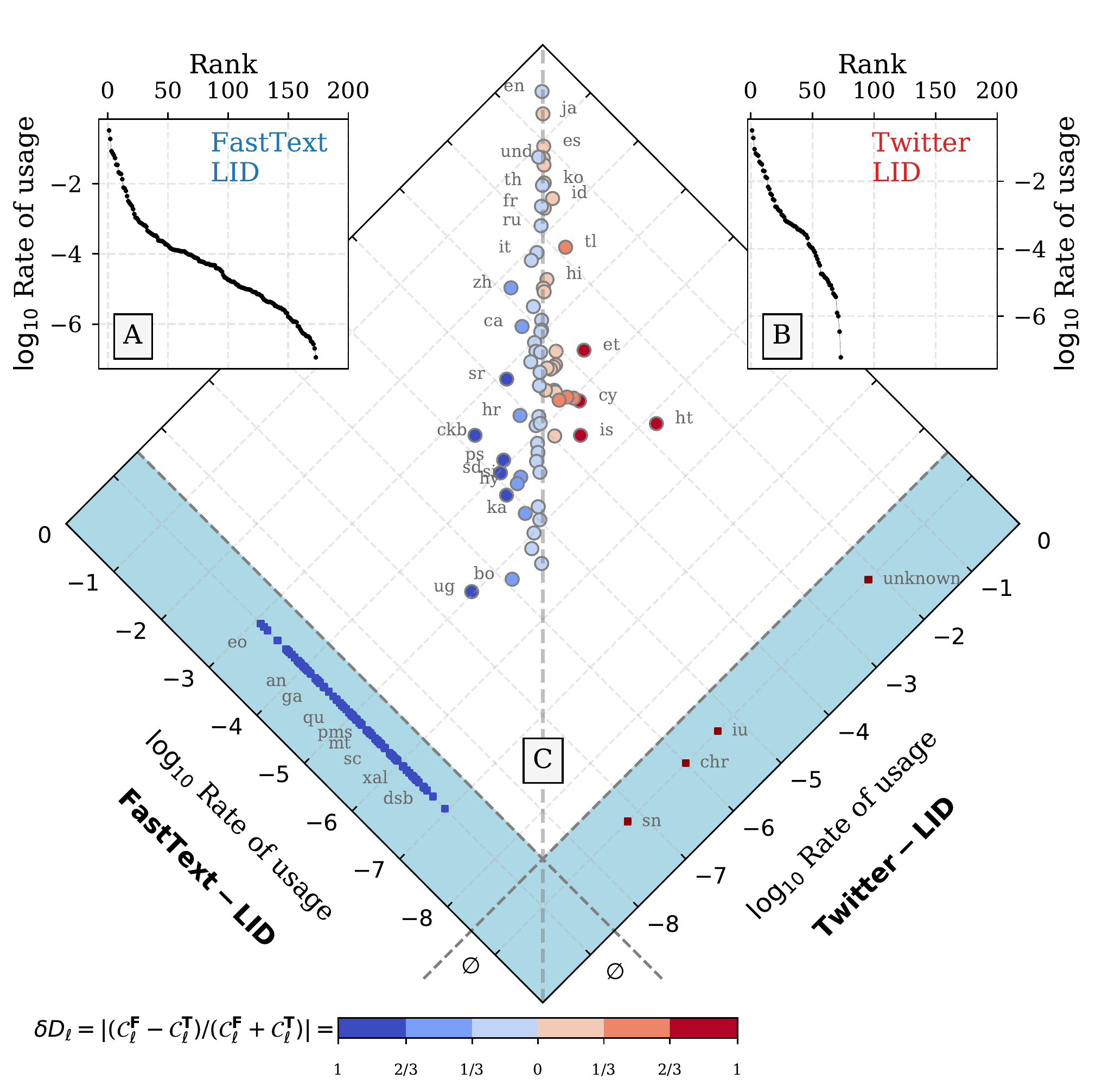}
\caption{
\textbf{Language Zipf distributions.}
\textbf{A.} 
Zipf distribution~\cite{zipf1949HumanBA} of all languages captured by FastText-LID model. 
\textbf{B.} 
Zipf distribution for languages captured by Twitter-LID algorithm(s).
The vertical axis in both panels reports rate of usage of all messages $p_{t,\ell}$
between 2014 and 2019, 
while the horizontal axis shows the corresponding rank of each language. 
FastText-LID recorded a total of 173 unique languages throughout that period.
On the other hand, 
Twittert-LID captured a total of 73 unique languages throughout that same period, 
some of which were experimental and no longer available in recent years. 
\textbf{C.} Joint distribution of all recorded languages. 
Languages located near the vertical dashed gray line 
signify agreement between FastText-LID and Twitter-LID, 
specifically that they 
captured a similar number of messages between 2014 and end of 2019.  
Languages found left of this line are more prominent using the FastText-LID model,
whereas languages right of the line are identified more frequently by Twitter-LID model.
Languages found within the light-blue area are only detectable by one classifier 
but not the other where FastText-LID is colored in blue and Twitter is colored in red.
The color of the points highlights the normalized ratio difference 
$\delta D_{\ell}$ (i.e., divergence)
between the two classifiers, 
where 
$\mathcal{C}^{\textnormal{F}}_{\ell}$ 
is the number of messages captured by FastText-LID for language ${\ell}$,
and 
$\mathcal{C}^{\textnormal{T}}_{\ell}$ 
is the number of messages captured by Twitter-LID for language ${\ell}$.
Hence, 
points with darker colors indicate greater divergence between the two classifiers.
A lookup table for language labels can be found in the Table~\ref{tab:iso-codes}, 
and an online appendix of all languages is also available here: 
\href{http://compstorylab.org/storywrangler/papers/tlid/files/fasttext\_twitter\_timeseries.html}{http://compstorylab.org/storywrangler/papers/tlid/files/fasttext\_twitter\_timeseries.html}.
} 
\label{fig:divergence}
\end{figure*}

We observe some disagreement between the two classifiers during the early years 
of Twitter's introduction of the LID feature to the platform.
In Fig.~\ref{fig:divergence}, 
we show the normalized ratio difference $\delta D_{\ell}$ (i.e., divergence)
between the two classifiers for all messages between 2014 and 2019.
Divergence is calculated as:
\begin{align}\label{divergence}
    \delta D_{\ell} = \bigg| \dfrac{\mathcal{C}^{\textnormal{F}}_{\ell} - \mathcal{C}^{\textnormal{T}}_{\ell}}{\mathcal{C}^{\textnormal{F}}_{\ell} + \mathcal{C}^{\textnormal{T}}_{\ell}} \bigg|,
\end{align}
where $\mathcal{C}^{\textnormal{F}}_{\ell}$ is the number of messages captured by FastText-LID for language ${\ell}$,
and $\mathcal{C}^{\textnormal{T}}_{\ell}$ is the number of messages captured by Twitter-LID for language ${\ell}$. 

\begin{figure*}[tp!] 
\includegraphics[width=\textwidth]{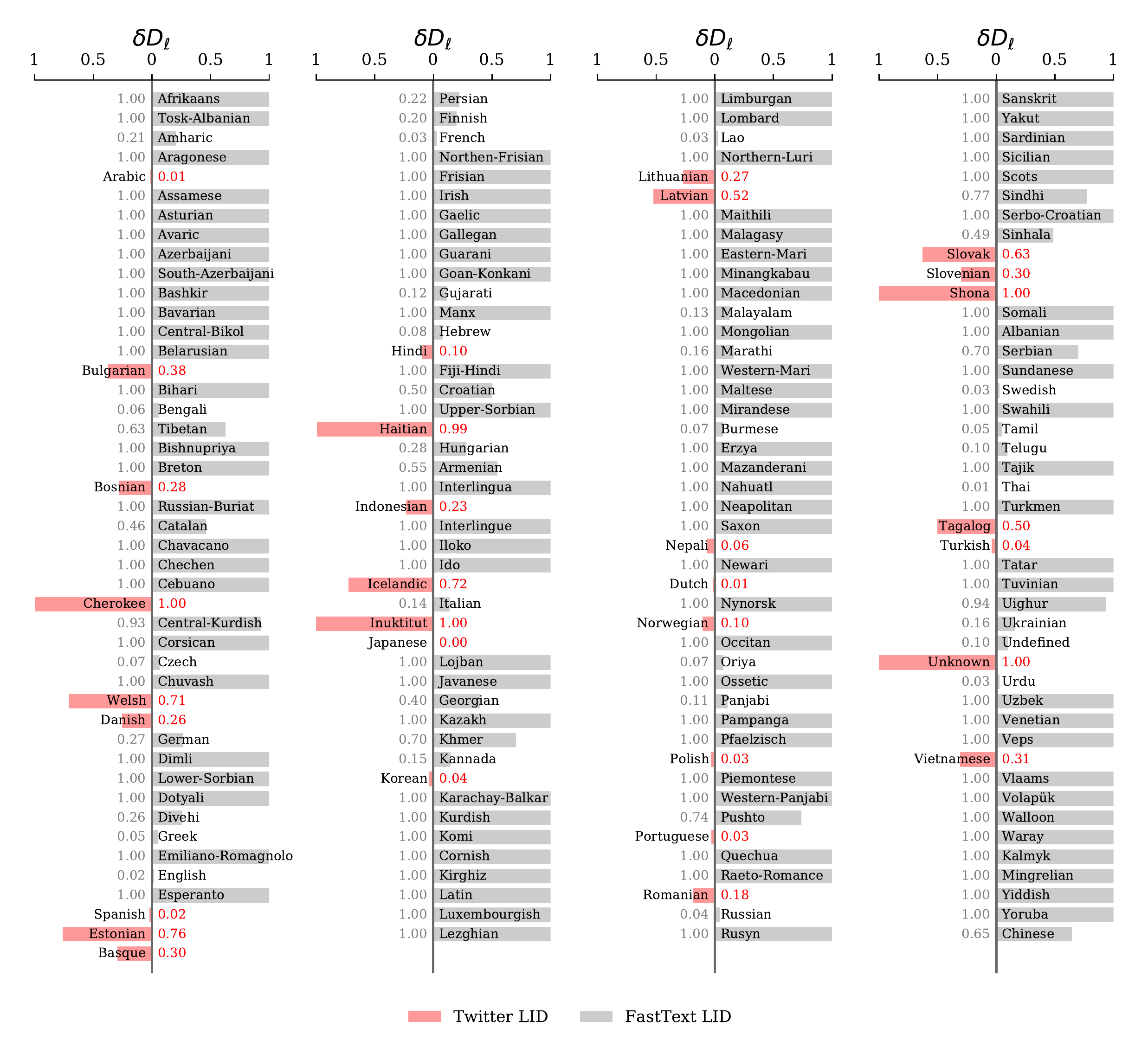}  
\caption{ 
\textbf{Language identification divergence.}
A normalized ratio difference value $\delta D_{\ell}$ (i.e., divergence)
closer to zero implies strong agreement,
whereby both classifiers captured approximately 
the same number of messages over the last decade. 
Grey bars indicate higher rate of messages captured by FastText-LID, 
whereas red bars highlight higher rate of messages captured by Twitter-LID. 
}
\label{fig:shifts} 
\end{figure*}

We show Zipf distributions of all languages captured by FastText-LID and
Twitter-LID in Fig.~\ref{fig:divergence}A and Fig.~\ref{fig:divergence}B, respectively.
FastText-LID recorded a total of 173 unique languages, 
whereas Twitter-LID captured a total of 73 unique languages throughout that period. 
Some of the languages reported by Twitter were experimental and no longer available in recent years. 
In Fig.~\ref{fig:divergence}C, 
we display the joint distribution of all languages captured by both classifiers.
Languages found left of vertical dashed gray line are more prominent using the FastText-LID model 
(e.g., Chinese (zh), Central-Kurdish (ckb), Uighur (ug), Sindhi (sd)).
Languages right of the line are identified more frequently by the Twitter-LID model 
(e.g., Estonian (et), Haitian (ht)).
Languages found within the light-blue area are only detectable by one classifier but not the other.
We note that `Unknown' is an artificial label that we added to flag messages
with missing language labels in the metadata of our dataset. 
We list divergence values $\delta D_{\ell}$ for all languages identified in our dataset in Fig.~\ref{fig:shifts}.

\clearpage
\section{Analytical validation of contagion ratios}
\label{sec:validation}

\begin{table*}[h!]
\centering
\caption{Top 10 languages with the highest annual average contagion ratio (sorted by 2019).}
\begin{tabular}{l|rrrrrrrrrrr}
\toprule \textbf{Language} &  \textbf{2009} &  \textbf{2010} &  \textbf{2011} &  \textbf{2012} &  \textbf{2013} &  \textbf{2014} &  \textbf{2015} &  \textbf{2016} &  \textbf{2017} &  \textbf{2018} &  \textbf{2019} \\
\hline\hline\midrule
Greek    &  0.01 &  0.05 &  0.07 &  0.20 &  0.42 &  0.65 &  0.83 &  1.11 &  1.29 &  1.42 &  1.27 \\
French   &  0.02 &  0.10 &  0.13 &  0.22 &  0.34 &  0.56 &  0.76 &  0.94 &  1.09 &  1.40 &  1.37 \\
English  &  0.03 &  0.14 &  0.20 &  0.31 &  0.37 &  0.56 &  0.71 &  0.91 &  1.15 &  1.44 &  1.44 \\
Spanish  &  0.03 &  0.16 &  0.21 &  0.31 &  0.42 &  0.62 &  0.82 &  0.94 &  1.24 &  1.54 &  1.52 \\
Korean   &  0.05 &  0.11 &  0.14 &  0.26 &  0.30 &  0.43 &  0.66 &  1.28 &  1.74 &  2.22 &  2.07 \\
Catalan  &  0.01 &  0.08 &  0.12 &  0.21 &  0.30 &  0.52 &  0.74 &  0.98 &  1.80 &  2.44 &  2.10 \\
Urdu     &  0.03 &  0.25 &  0.25 &  0.19 &  0.26 &  0.64 &  0.82 &  0.95 &  1.51 &  2.67 &  2.90 \\
Tamil    &  0.01 &  0.04 &  0.10 &  0.16 &  0.22 &  0.54 &  0.82 &  1.30 &  1.84 &  2.40 &  2.96 \\
Hindi    &  0.01 &  0.03 &  0.06 &  0.15 &  0.38 &  1.14 &  2.26 &  2.81 &  3.09 &  3.58 &  3.29 \\
Thai     &  0.07 &  0.24 &  0.18 &  0.32 &  0.79 &  2.01 &  2.54 &  3.35 &  5.31 &  6.52 &  7.29 \\
\end{tabular}
\label{tab:top_ratios}
\end{table*}

\begin{table*}[h!]
\centering
\caption{Bottom 10 languages with the lowest annual average contagion ratio (sorted by 2019).}
\begin{tabular}{l|rrrrrrrrrrr}
\toprule \textbf{Language} &  \textbf{2009} &  \textbf{2010} &  \textbf{2011} &  \textbf{2012} &  \textbf{2013} &  \textbf{2014} &  \textbf{2015} &  \textbf{2016} &  \textbf{2017} &  \textbf{2018} &  \textbf{2019} \\
\hline\hline\midrule
Finnish   &  0.02 &  0.11 &  0.10 &  0.11 &  0.14 &  0.18 &  0.23 &  0.26 &  0.29 &  0.31 &  0.26 \\
Cebuano   &  0.01 &  0.07 &  0.09 &  0.13 &  0.14 &  0.22 &  0.24 &  0.29 &  0.32 &  0.33 &  0.30 \\
Esperanto &  0.01 &  0.08 &  0.09 &  0.11 &  0.13 &  0.18 &  0.24 &  0.34 &  0.41 &  0.47 &  0.38 \\
Swedish   &  0.02 &  0.07 &  0.09 &  0.14 &  0.20 &  0.31 &  0.37 &  0.41 &  0.47 &  0.55 &  0.45 \\
Russian   &  0.01 &  0.04 &  0.07 &  0.13 &  0.13 &  0.19 &  0.29 &  0.31 &  0.42 &  0.57 &  0.50 \\
Dutch     &  0.02 &  0.11 &  0.16 &  0.23 &  0.23 &  0.28 &  0.32 &  0.36 &  0.42 &  0.52 &  0.51 \\
German    &  0.02 &  0.07 &  0.09 &  0.13 &  0.17 &  0.26 &  0.34 &  0.38 &  0.42 &  0.58 &  0.52 \\
Japanese  &  0.02 &  0.08 &  0.10 &  0.11 &  0.16 &  0.31 &  0.35 &  0.31 &  0.40 &  0.47 &  0.53 \\
Polish    &  0.01 &  0.06 &  0.08 &  0.13 &  0.22 &  0.28 &  0.42 &  0.60 &  0.84 &  0.74 &  0.57 \\
Persian   &  0.03 &  0.07 &  0.07 &  0.14 &  0.22 &  0.40 &  0.35 &  0.41 &  0.50 &  0.64 &  0.57 \\
\end{tabular}
\label{tab:bottom_ratios}
\end{table*}

To investigate our margin of error for estimating contagion ratios, 
we find the subset of messages that both classifiers have agreed on their language labels
using the annual confusion matrices we discussed in Appendix~\ref{sec:Decahose}. 
We compute an annual average of the contagion ratios for this subset of messages.
We highlight the top 10 languages with the highest and lowest average contagion ratio per year in Table~\ref{tab:top_ratios} and Table~\ref{tab:bottom_ratios}, respectively.
We then compare the new set of annual contagion ratios with the original ones discussed in Sec.~\ref{sec:ratio}.
In particular, 
we compute the absolute difference 
$$\delta = |\textnormal{R} - \textnormal{R}_{\alpha}|,$$
where $\textnormal{R}$ indicates the contagion ratios of all messages, 
and $\textnormal{R}_{\alpha}$ indicates the contagion ratios 
of the subset of messages that both FastText-LID and Twitter-LID models have unanimously agreed on their language labels.

In Table~\ref{tab:top_errors}, 
we show the top 10 languages with the highest average values of $\delta$'s.
Languages are sorted by the values of $\delta$'s in 2019. 
Higher values of $\delta$'s indicate high uncertainty 
due to high disagreement on the language of the written messages 
between FastText-LID and Twitter-LID. 
Lower values of $\delta$'s, on the other hand, 
highlight better agreement between the two classifiers,
and thus better confidence in our estimation of the contagion ratios.
We show the bottom 10 languages with the lowest average values of $\delta$'s in Table~\ref{tab:bottom_errors}.

\clearpage
\begin{table}[h!]
\begin{minipage}[t]{0.47\linewidth}
\centering
\caption{Top 10 languages with the highest average margin of error 
for estimating contagion ratios as a function 
of the agreement between FastText-LID and Twitter-LID (sorted by 2019).}
\begin{tabular}{l|rrrrrr}
\toprule \textbf{Language} & \textbf{2014} &  \textbf{2015} &  \textbf{2016} &  \textbf{2017} &  \textbf{2018} &  \textbf{2019} \\
\hline\hline\midrule
Undefined  &  $\pm$0.14 &  $\pm$0.14 &  $\pm$0.16 &  $\pm$0.19 &  $\pm$0.17 &  $\pm$0.15 \\
Dutch      &  $\pm$0.11 &  $\pm$0.10 &  $\pm$0.11 &  $\pm$0.12 &  $\pm$0.15 &  $\pm$0.17 \\
Swedish    &  $\pm$0.14 &  $\pm$0.16 &  $\pm$0.18 &  $\pm$0.19 &  $\pm$0.21 &  $\pm$0.20 \\
Serbian    &  $\pm$0.26 &  $\pm$0.27 &  $\pm$0.32 &  $\pm$0.33 &  $\pm$0.35 &  $\pm$0.25 \\
Cebuano    &  $\pm$0.22 &  $\pm$0.24 &  $\pm$0.29 &  $\pm$0.32 &  $\pm$0.33 &  $\pm$0.30 \\
Esperanto  &  $\pm$0.18 &  $\pm$0.24 &  $\pm$0.34 &  $\pm$0.41 &  $\pm$0.47 &  $\pm$0.38 \\
Indonesian &  $\pm$0.21 &  $\pm$0.18 &  $\pm$0.18 &  $\pm$0.24 &  $\pm$0.39 &  $\pm$0.40 \\
Tagalog    &  $\pm$0.22 &  $\pm$0.34 &  $\pm$0.49 &  $\pm$0.51 &  $\pm$0.48 &  $\pm$0.44 \\
Hindi      &  $\pm$0.08 &  $\pm$0.41 &  $\pm$0.97 &  $\pm$0.76 &  $\pm$0.73 &  $\pm$0.71 \\
Catalan    &  $\pm$0.52 &  $\pm$0.74 &  $\pm$0.98 &  $\pm$1.80 &  $\pm$1.08 &  $\pm$0.75 \\ 
\end{tabular}
\label{tab:top_errors}
\end{minipage}\hfill%
\begin{minipage}[t]{0.47\linewidth}
\centering
\caption{Bottom 10 languages with the lowest average margin of error 
for estimating contagion ratios as a function 
of the agreement between FastText-LID and Twitter-LID (sorted by 2019).}
\begin{tabular}{l|rrrrrr}
\toprule \textbf{Language} & \textbf{2014} &  \textbf{2015} &  \textbf{2016} &  \textbf{2017} &  \textbf{2018} &  \textbf{2019} \\
\hline\hline\midrule
Tamil      &  $\pm$0.03 &  $\pm$0.01 &  $\pm$0.01 &  $\pm$0.01 &  $\pm$0.01 &  $\pm$0.01 \\
Greek      &  $\pm$0.13 &  $\pm$0.07 &  $\pm$0.01 &  $\pm$0.01 &  $\pm$0.01 &  $\pm$0.01 \\
Japanese   &  $\pm$0.01 &  $\pm$0.01 &  $\pm$0.01 &  $\pm$0.01 &  $\pm$0.02 &  $\pm$0.02 \\
Russian    &  $\pm$0.01 &  $\pm$0.01 &  $\pm$0.01 &  $\pm$0.02 &  $\pm$0.03 &  $\pm$0.03 \\
Persian    &  $\pm$0.10 &  $\pm$0.06 &  $\pm$0.06 &  $\pm$0.05 &  $\pm$0.04 &  $\pm$0.03 \\
Arabic     &  $\pm$0.04 &  $\pm$0.03 &  $\pm$0.02 &  $\pm$0.02 &  $\pm$0.03 &  $\pm$0.04 \\
Chinese    &  $\pm$0.04 &  $\pm$0.04 &  $\pm$0.04 &  $\pm$0.05 &  $\pm$0.06 &  $\pm$0.08 \\
English    &  $\pm$0.04 &  $\pm$0.05 &  $\pm$0.05 &  $\pm$0.06 &  $\pm$0.08 &  $\pm$0.09 \\
Thai       &  $\pm$0.03 &  $\pm$0.03 &  $\pm$0.04 &  $\pm$0.06 &  $\pm$0.08 &  $\pm$0.09 \\
Portuguese &  $\pm$0.08 &  $\pm$0.10 &  $\pm$0.09 &  $\pm$0.11 &  $\pm$0.11 &  $\pm$0.10 \\
\end{tabular}
\label{tab:bottom_errors}
\end{minipage}
\end{table}

\begin{figure}[tp!] 
\includegraphics[width=\columnwidth]{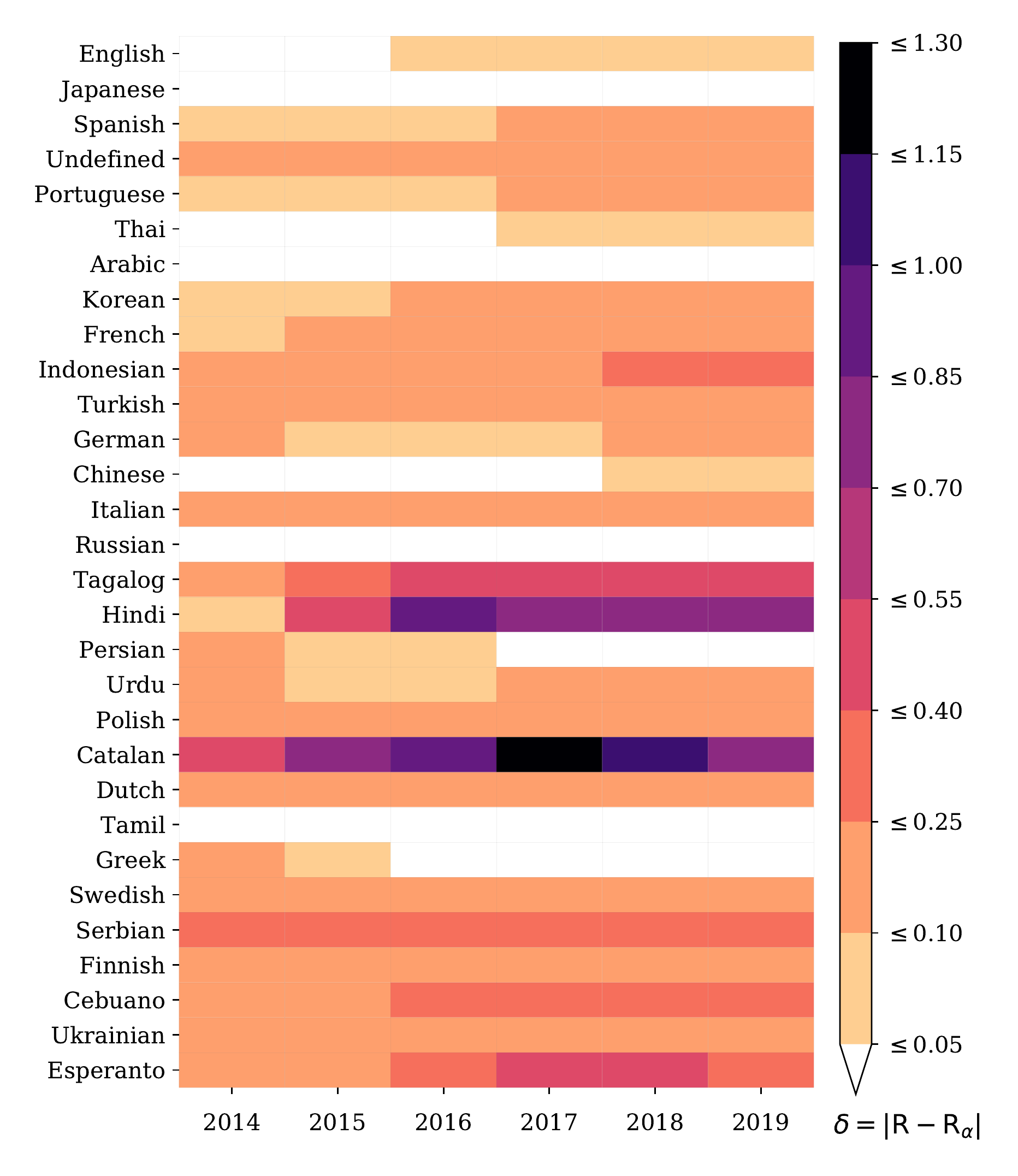} 
\caption{
\textbf{Margin of error for contagion ratios.}
We compute the annual average of contagion ratios $\textnormal{R}$ 
for all messages in the top 30 ranked languages as classified by FastText-LID and described in Sec.~\ref{sec:ratio}. 
Similarly, we compute the annual average of contagion ratios $\textnormal{R}_{\alpha}$
for the subset of messages that both classifiers have unanimously labeled their language labels.
We display the absolute difference $\delta = |\textnormal{R} - \textnormal{R}_{\alpha}|$
to indicate our margin of error for estimating contagion ratios as a function 
of the agreement between FastText-LID and Twitter-LID models. 
White cells indicate that $\delta$ is below $.05$,
whereas colored cells highlight values that are equal to, or above $.05$. 
We show the top 10 languages with the highest average values of $\delta$'s per year in Table~\ref{tab:top_errors}.
We also show the bottom 10 languages with the lowest average values of $\delta$'s per year in Table~\ref{tab:bottom_errors}.
}   
\label{fig:val_heatmap}    
\end{figure}

In Fig.~\ref{fig:val_heatmap},
we display a heatmap of $\delta$'s for the top 30 ranked languages. 
We note low values of $\delta$'s for the top 10 languages on the platform. 
In other words, our contagion ratios for the subset of messages 
that both classifiers have unanimously predicted their language labels are roughly equivalent to our estimations in Table~\ref{tab:top_ratios}.   
By contrast, 
we note high disagreement on Catalan messages.  
The two classifiers start off with unusual disagreement in 2014 ($\delta = .52$). 
The disagreement between the two models continues to grow 
leading to a remarkably high value of $\delta =1.80$ in 2017.
Thereafter, we observe a trend down in our estimations, indicating that FastText-LID and Twitter-LID have slowly started to harmonize their language predictions for Catalan messages through the past few years.  
We also note similar trends for Hindi and Tagalog messages. 

Our results show empirical evidence of inconsistent language labels in the historical data feed between 2014 and 2017. 
Our margin of error for estimating contagion ratios narrows down as FastText-LID and Twitter-LID unanimously yield their language predictions for the majority of messages authored in recent years.
Future investigations can help us shed light on some of the implicit biases of language detection models. 
Nonetheless, our analysis supports our findings regarding the growth of retweets over time across most languages.

\clearpage
\section{Impact of tweet's length on language detection}
\label{sec:tweet_len}

The informal and short texture of messages on 
Twitter---among many other reasons---makes language detection of tweets remarkably challenging. 
Twitter has also introduced several changes to the platform that notably impacted language identification. 
Particularly, users were limited to 140 characters per message before the last few months of 2017
and 280 characters thereafter~\cite{tweet_len}.
To investigate the level of uncertainty of language detection as a function of tweet length, we take a closer look at the number of messages that are classified differently by FastText-LID and Twitter-LID for the top 10 most used languages on the platform
between 2020-01-01 and 2020-01-07.

\begin{figure}[h!] 
\includegraphics[width=.98\columnwidth]{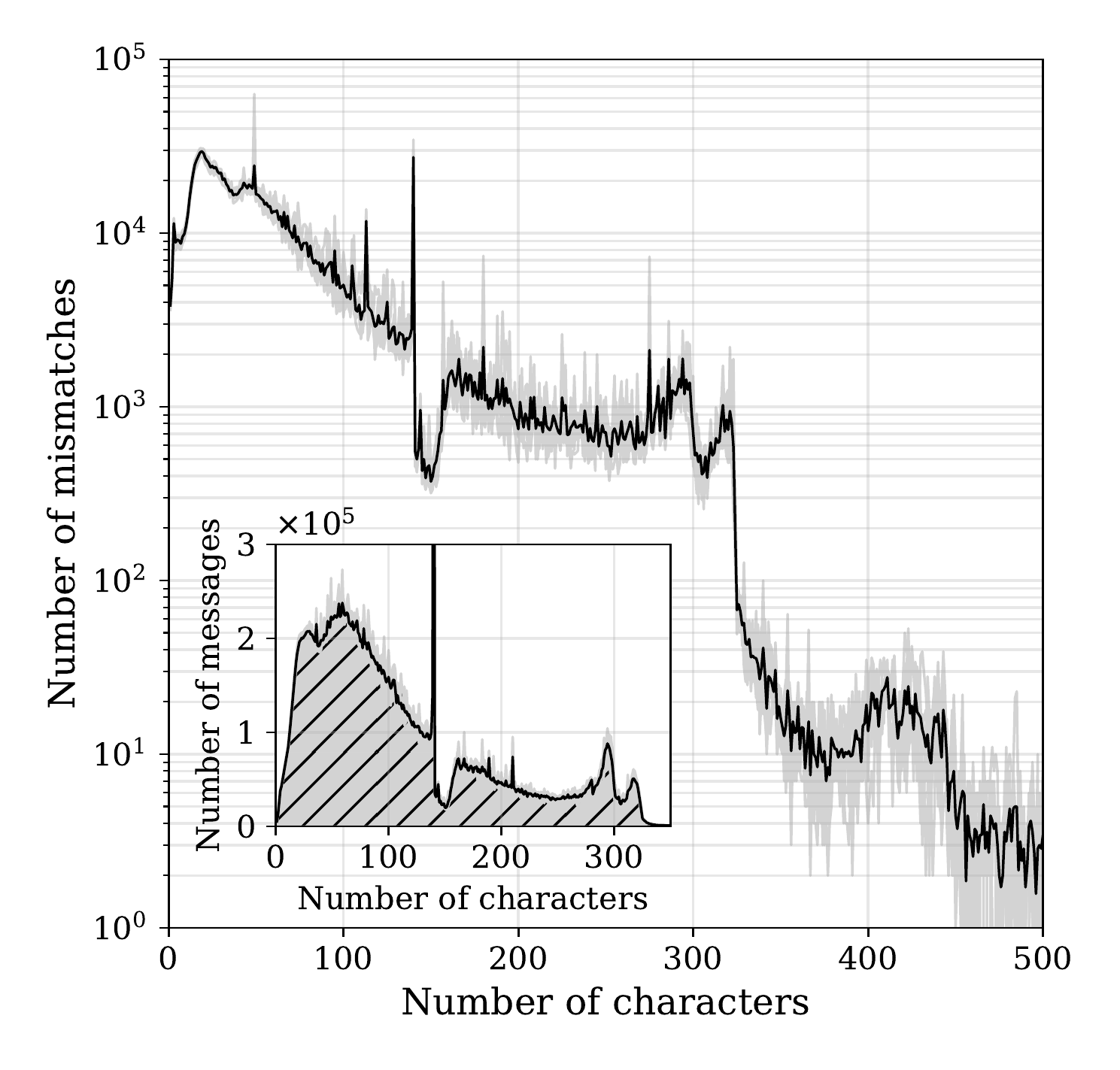} 
\caption{
\textbf{Language identification uncertainty as a function of tweet-length for top 10 most used languages on Twitter.}
We display the number of messages that were classified differently
by Twitter-LID model and FastText-LID for the top-10 prominent languages 
as a function of the number of characters in each message.
Unlike Twitter, we count each character individually,
which is why the length of each message may exceed the 280 character limit.   
The grey lines indicate the daily number of mismatches between 2020-01-01 and 2020-01-07
(approximately $32$ million messages for each day for the top-10 used languages),
whereas the black line shows an average of that whole week. 
} 
\label{fig:mismatches} 
\end{figure} 

\begin{table}[h!]
\centering
\caption{Average daily messages for the top 10 languages between 2020-01-01 and 2020-01-07 
(approximately $32$ million messages for each day).}
\begin{tabular}{l|cc}
\toprule \textbf{Language} & \textbf{Messages} & \textbf{Mismatches} \\
\hline\hline\midrule
English     & 1.1   $\times 10^7$ & .0853  \\
Japanese    & 6.8   $\times 10^6$ & .0268  \\
Spanish     & 2.3   $\times 10^6$ & .0558  \\
Thai        & 2.2   $\times 10^6$ & .0161  \\
Portuguese  & 2.1   $\times 10^6$ & .0565  \\
Korean      & 1.7   $\times 10^6$ & .0085  \\
Arabic      & 1.5   $\times 10^6$ & .0080  \\
Indonesian  & 8.1   $\times 10^5$ & .1203  \\
French      & 7.9   $\times 10^5$ & .1305  \\
Turkish     & 5.6   $\times 10^5$ & .0325  \\
\end{tabular}
\label{tab:top_mismatches}
\end{table}

In Fig.~\ref{fig:mismatches}, 
we display the daily number of mismatches (grey bars) between 2020-01-01 and 2020-01-07 
(approximately 32 million messages for each day for the top-10 used languages), 
whereas the black line shows an average of that whole week. 
We also display a histogram of the average number of characters of each message throughout that period.
We note that the distribution is remarkably skewed towards shorter messages on the platform.  
The average length of messages is less than 140 characters, with a large spike around the 140 character mark.  
Long messages---which include messages with links, hashtags, 
and emojis---can exceed the theoretical 280 character limit because 
we do not follow the same guidelines outlined by Twitter for counting the number of characters in each 
message.\footnote{\url{https://developer.twitter.com/en/docs/basics/counting-characters}}
For simplicity, we use the built-in Python function
to get the exact number of characters in a given message.\footnote{\url{https://docs.python.org/3/library/functions.html\#len}}
As anticipated, our results indicate a higher proportion of short messages classified differently by FastText-LID and Twitter-LID models. 
We highlight the average percentage of mismatches for the top 10 most used languages in Table~\ref{tab:top_mismatches}
(languages are sorted by popularity).

\begin{figure}[tp!]  
\includegraphics[width=.8\columnwidth]{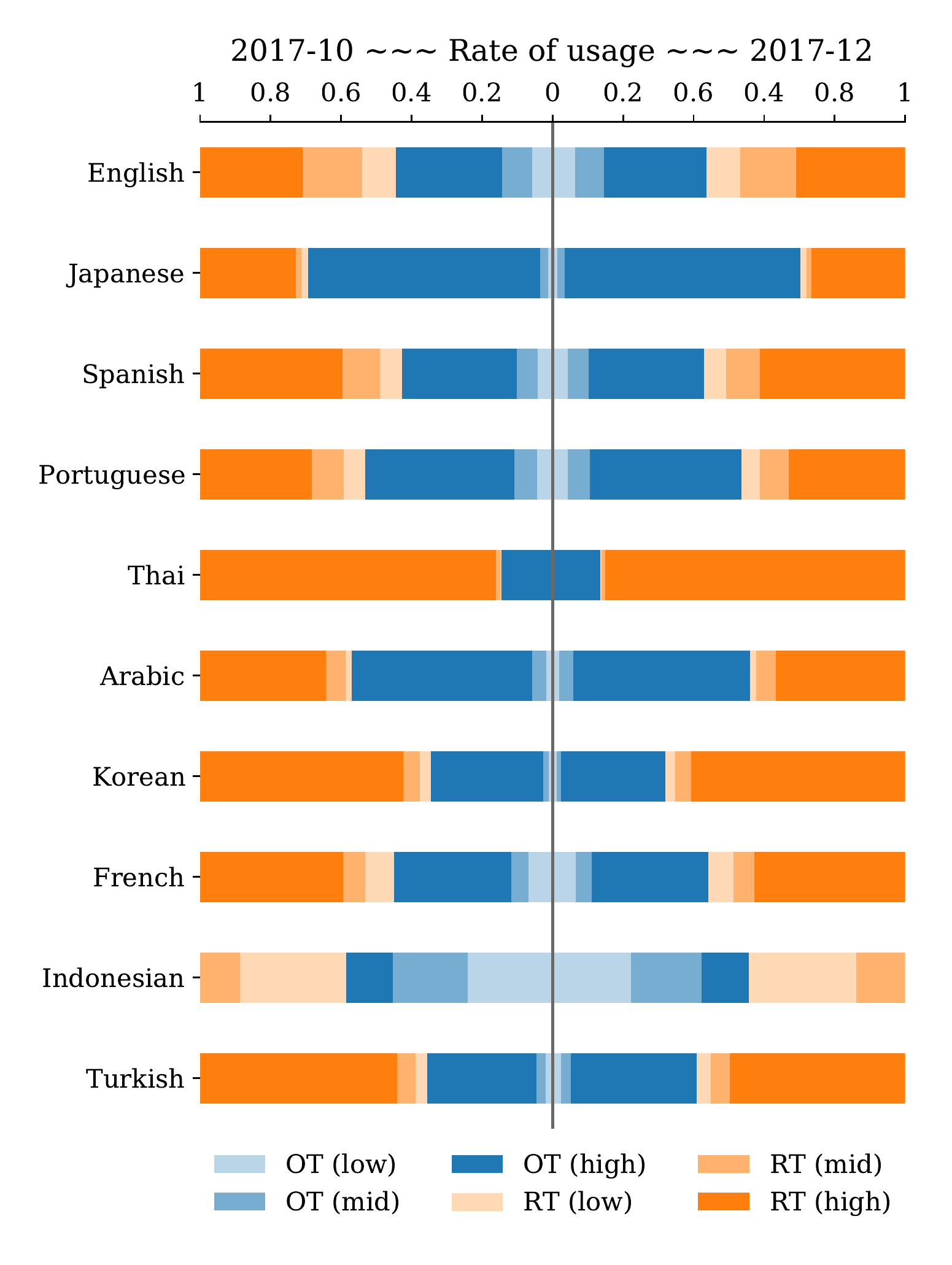} 
\caption{
\textbf{Confidence scores of the FastText-LID neural network predictions
for the month before and after the shift to 280 characters.}
We categorize messages into four classes based on the confidence scores 
we get from FastText-LID's neural network.
Predictions with confidence scores below .25 are labeled as Undefined (und).
Messages with scores greater or equal to .25 but less than .5 
are flagged as predictions with low confidence (low).
Predictions that have scores in the range [.5, .75) are considered moderate (mid),
and messages with higher scores are labeled as predictions with high confidence (high).
We note a symmetry indicating that the shift did not have 
a large impact on the network's predictions across organic and retweeted messages.
}   
\label{fig:agreement}   
\end{figure}

Furthermore, 
we examine a sample of messages authored through the month before and after the switch to the 280 character limit. 
We do not observe any distributional changes in FastText-LID's confidence scores between the two months. 
We categorize messages into four classes based on the confidence scores we get from FastText-LID's neural network.
Predictions with confidence scores below .25 are labeled as Undefined (und).
On the other hand, 
messages with scores greater or equal to .25 but less than .5 are flagged as predictions with low confidence (low).
Predictions that have scores in the range [.5, .75) are considered moderate (mid),
and messages with higher scores are labeled as predictions with high confidence (high).

In Fig.~\ref{fig:agreement}, 
we display the relative proportion of messages for each of the confidence classes outlined above.
First and foremost, we observe a very symmetrical layout indicating that the shift does not have a notable impact on the network's confidence in its predictions between the two months examined here
across organic and retweeted messages.

\begin{figure*}[th!] 
\includegraphics[width=\textwidth]{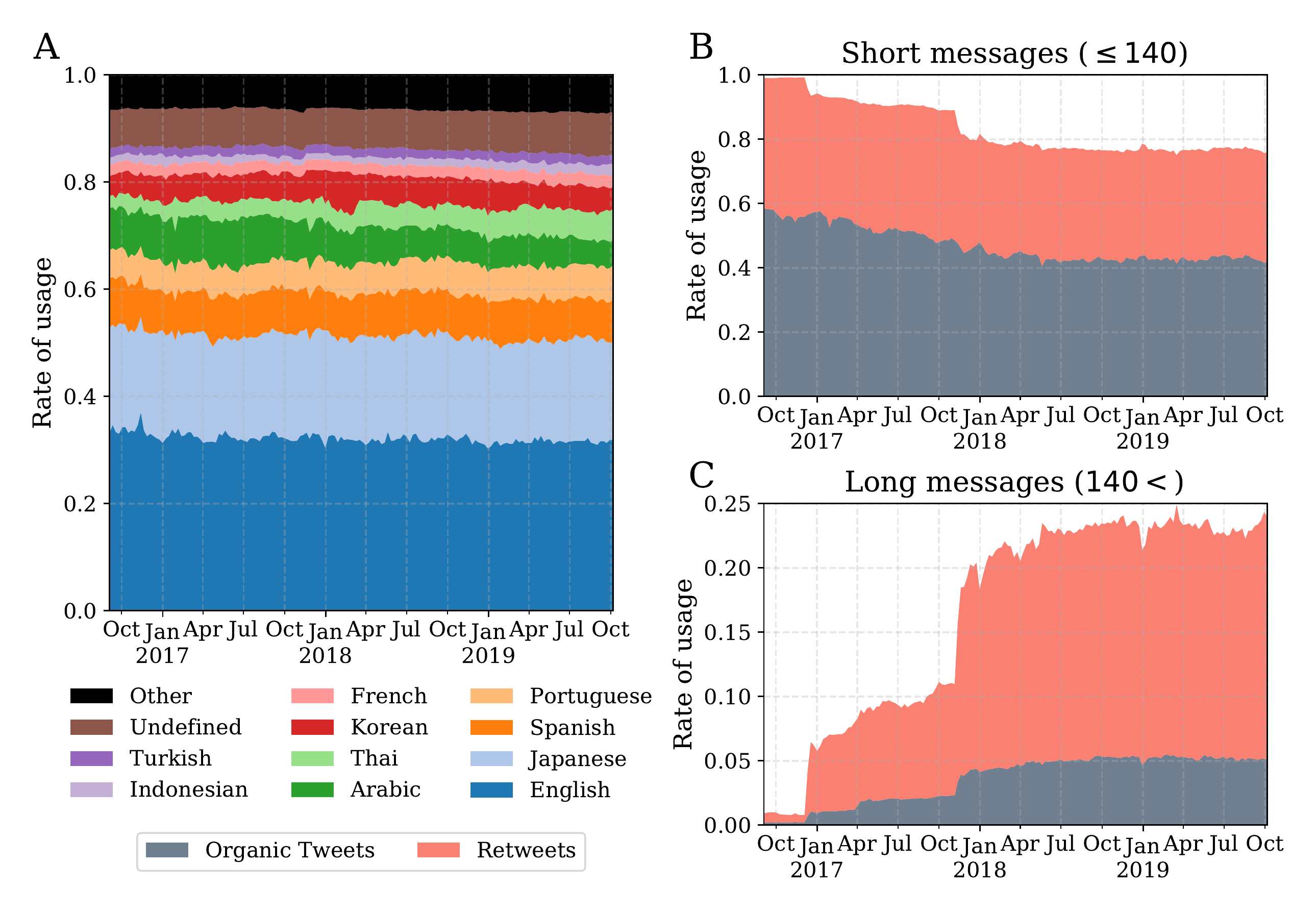} 
\caption{
\textbf{Weekly rate of usage for short and long messages.}
\textbf{A.} 
Rate of usage for the top-10 used languages averaged at the week scale for the past three years. 
The introduction of long messages 
(i.e., above 140 but below 280 characters)
does not change the overall language usage on the platform.  
\textbf{B--C.} 
The growth of long messages over time across organic and retweeted messages. 
We observe a much higher ratio of retweets in longer messages than shorter messages.
}   
\label{fig:usage}   
\end{figure*}

Moreover, 
we note that the overall rate of usage for each language does not change 
before and after the switch to longer messages. 
To validate that, 
we take a closer look at the rate of usage for the top 10 most used languages 
throughout the past three years. 
In Fig.~\ref{fig:usage}A, 
we observe a very consistent frequency of usage across all languages,
indicating that the mechanistic shift to allow users to post longer messages 
does not have a notable impact on the language detection process.
Fig.~\ref{fig:usage}B and Fig.~\ref{fig:usage}C 
show the growth of long messages on the platform,
while the rate of usage for the most used languages remains consistent.
In Fig.~\ref{fig:usage}C, 
we see the adoption of longer messages starting in 2017,
however, short messages still represent the majority of messages on the platform
which comprise 75\% of all messages as of 2019. 

We observe a much higher ratio of retweets in longer messages than shorter messages. 
As of 2019, about 25\% of all messages are long messages, and
surprisingly, 80\% of these long messages are retweets. 
However, we only examined the use of languages over time from a statistical point of view. 
The use of longer messages and the rate at which they are likely to be retweeted are different across languages. 
Further investigations will be needed to explore and explain this phenomenon.

\end{document}